\pdfoutput=1

\documentclass[11pt]{article}

\usepackage[final]{acl}

\usepackage{times}
\usepackage{latexsym}
\usepackage{graphicx}
\usepackage{array}
\usepackage{makecell}
\usepackage{bm}
\usepackage{multirow}
\usepackage{amsfonts}
\usepackage{colortbl} 
\usepackage{booktabs}
\usepackage{url}

\usepackage[most]{tcolorbox} 

\definecolor{navy}{RGB}{50,110,168}

\usepackage{hyperref}

\usepackage[T1]{fontenc}

\usepackage[utf8]{inputenc}

\usepackage{microtype}

\usepackage{enumitem}

\usepackage{inconsolata}

\newcommand{\graycell}{\cellcolor[rgb]{0.957,0.957,0.957}}

\newcommand{\grayrow}{\rowcolor[rgb]{0.957,0.957,0.957}}
\newcommand{\bluerow}{\rowcolor[rgb]{0.907,0.977,0.957}}

\usepackage{xspace}
\newcommand{\CoReTab}{{\texttt{CoReTab}}\xspace}
\newcommand{\code}{\textbf{\texttt{<code>}}\xspace}
\newcommand{\answer}{\textbf{\texttt{<answer>}}\xspace}
\newcommand{\reason}{\textbf{\texttt{<reason>}}\xspace}

\author{Van-Quang Nguyen \\
  RIKEN AIP, Japan \\
  \texttt{quang.nguyen.jz@riken.jp} \\\And
  Takayuki Okatani \\
  GSIS, Tohoku University/RIKEN AIP, Japan \\
  \texttt{okatani@vision.is.tohoku.ac.jp} \\}

\title{\CoReTab: Improving Multimodal Table Understanding with Code-driven Reasoning}  

\begin{document}
\maketitle
\begin{abstract}
  Existing datasets for multimodal table understanding, such as MMTab, primarily provide short factual answers without explicit multi-step reasoning supervision. Models trained on these datasets often generate brief responses that offers insufficient accuracy and limited interpretability into how these models arrive at the final answer. We introduce \textbf{\CoReTab}, a code-driven reasoning framework that produces scalable, interpretable, and automatically verifiable annotations by coupling multi-step reasoning with executable Python code. Using the \CoReTab framework, we curate a dataset of 115K verified samples averaging 529 tokens per response and fine-tune open-source MLLMs through a three-stage pipeline.
  We evaluate the resulting model trained on \CoReTab across 17 MMTab benchmarks spanning table question answering, fact verification, and table structure understanding. Our model achieves significant gains of +6.2\%, +5.7\%, and +25.6\%, respectively, over MMTab-trained baselines, while producing transparent and verifiable reasoning traces. These results establish \CoReTab as a robust and generalizable supervision framework for improving multi-step reasoning in multimodal table understanding.
\end{abstract}

\begin{figure*}[t]
  \centering
  \includegraphics[width=2.\columnwidth]{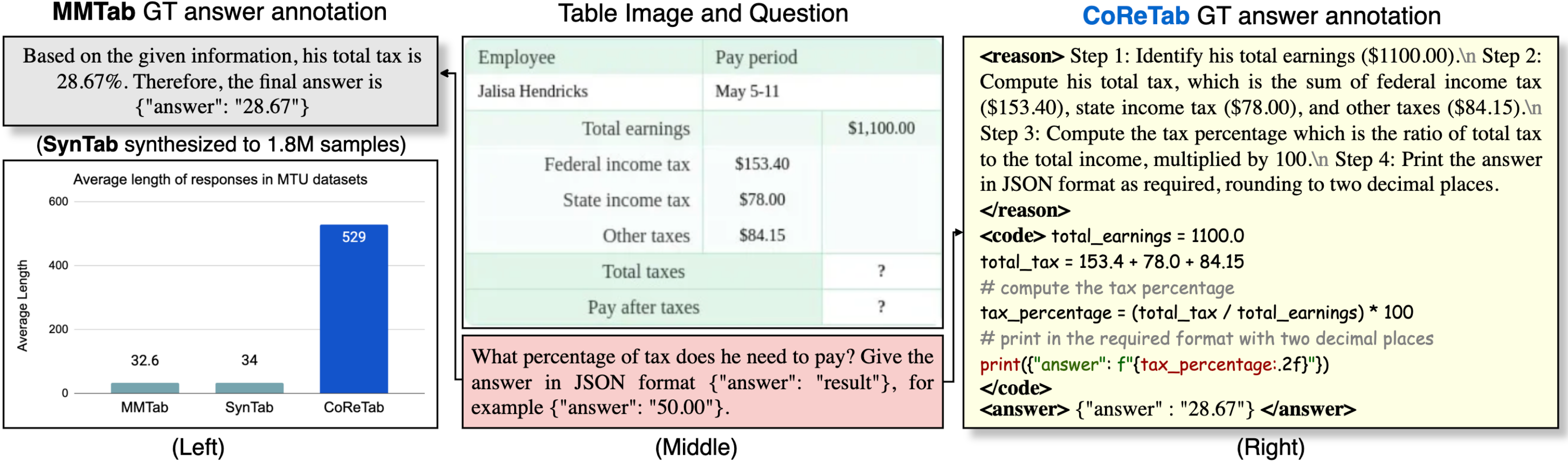}
  \caption{(Left-top) Typical response annotation from MMTab \cite{MMTab} which provides limited reasoning supervision. (Left-bottom) Average length of responses of three datasets.
  (Middle) A pair of ⟨table image, question⟩.
  (Right) Our {\bf\CoReTab} annotation example with step-by-step reasoning and executable Python code.}
  \label{fig:intro}  
\end{figure*}

\section{Introduction}
  Tables serve as a primary medium for organizing and communicating (semi-)structured information, with applications spanning finance, science, government, and diverse practical domains \citep{web_table_taxonomy,tu_survey_2023}. In real-world settings, tables are often embedded as screenshots in documents such as slides, PDFs, or spreadsheets, making the table inherently a visual object. Text-only large language models (LLMs) are ill-suited for such inputs, as extracting text can disrupt spatial layout and introduce inconsistencies. Moreover, directly interacting with table images (e.g., cropping or highlighting) is often more natural and efficient than relying on OCR followed by complex textual representations such as HTML or LaTeX. Prior work shows that even OCR+LLM pipelines substantially underperform multimodal large language models (MLLMs) on table image understanding \citep{MMTab}. Consequently, growing attention has focused on multimodal table understanding, which aims to enable machines to comprehend and reason over table images \citep{mmcoqa,luo2023unifying,MMTab,SynTab}.  
  Recent advances in multimodal large language models \citep{alayrac2022flamingo,openai2023gpt4v,bai2025qwen2} have moved us closer to general-purpose multimodal table understanding, yet their performance remains constrained.
  
  A central bottleneck lies in supervision. Human annotation, as in \citep{TabFact,HiTab,WTQ}, requires annotators to generate question–answer pairs for each table image, making it accurate but prohibitively costly and slow. Automated synthesis with MLLMs has been explored as an alternative \citep{zhao2024tabpedia,meng2024chartassisstant}, where the model itself is instructed to generate ⟨question, answer⟩ pairs, given table images. While scalable, this approach struggles with complex table images, resulting in hallucinations and reasoning errors.
  
  Another line of work leverages table codes (textual representations of tables in HTML, \LaTeX, or Markdown) that can be rendered into images and paired with generated annotations. 
  \citet{MMTab} curated MMTab, a large-scale multimodal dataset of 232K ⟨table image, question, answer⟩ samples by constructing from existing textual table datasets, partially employing a large language model to paraphrase the answers. \citet{SynTab} collected more table codes and scaled training resources to 1.8M samples using LLMs to generate question–answer pairs directly from table codes.
  The annotations in MMTab and SynTab are designed to primarily consist of short answers without multi-step reasoning (Figure~\ref{fig:intro}, left). Models trained on such supervision may excel at question answering but offer little transparency into how these models arrive at the final answer. For real-world applications, where correctness and interpretability are essential, this gap is critical.
  
  It is feasible to use (M)LLMs to automatically generate multi-step reasoning annotations; however, validating the correctness of lengthy reasoning text remains challenging and costly. We address this challenge with \textbf{\CoReTab}, a \textbf{Co}de-driven \textbf{Re}asoning framework for multimodal \textbf{Ta}ble understanding. \CoReTab couples natural-language multi-step reasoning with executable code, enabling automatic verification of reasoning correctness. In this framework, an LLM annotator (e.g., Qwen3~\citep{yang2025qwen3}) is prompted with a table code, a question, and a short ground-truth answer to produce a new annotation that includes both multi-step reasoning and corresponding Python code (Figure~\ref{fig:intro}, right).
  The Python code is executed and verified against the ground-truth answer, and only consistent samples are retained. Multi-step reasoning text associated with verified code tends to be more accurate and less prone to hallucination. As a result, this closed-loop, code-verifiable framework allows scalable generation of accurate, interpretable reasoning annotations with minimal human effort.

  Using this framework, we curate a \CoReTab dataset, an augmented version of MMTab containing 115K verified samples across 11 tasks, with average responses being on average an order of magnitude longer than those in MMTab or SynTab.
  Each annotation includes step-by-step reasoning and executable code, ensuring both interpretability and verifiability. Moreover, the process is cost-efficient: while annotating 2,000 examples for TableBench~\cite{wu2025tablebench} required an estimated \$12,000, \CoReTab scales to 115K samples at a tiny fraction of that cost using local GPU inference.

   To evaluate the effectiveness of \CoReTab, we fine-tune Qwen2.5-VL \citep{bai2025qwen2} using a three-stage training pipeline with LoRA \citep{hu2022lora}: (1) table recognition pretraining, (2) instruction tuning on \CoReTab, and (3) reinforcement learning optimization. The resulting model consistently outperforms MMTab-trained baselines (e.g., Table-Qwen2.5-VL) across 17 MMTab benchmarks, with average improvements of +6.2\% in table question answering, +5.7\% in table fact verification, and +25.6\% in table structure understanding. Overall, the code-driven reasoning supervision enhances robustness—by teaching models to couple reasoning with executable code—and interpretability, as each prediction provides multi-step reasoning that can be inspected and verified.
  
We summarize our contributions as follows:
\begin{itemize}\setlength\itemsep{0em}\setlength{\topsep}{0pt}
  \item We introduce \CoReTab, the first framework for code-driven reasoning supervision with automatic verification for multimodal table understanding.
  
  \item We curate a dataset of 115K verified annotations combining natural-language and executable reasoning, offering richer supervision than existing MMTab and SynTab datasets.

  \item We demonstrate that the resulting MLLM trained under \CoReTab supervision achieves substantial improvements across 17 MMTab benchmarks, delivering more accurate, interpretable, and verifiable answers.
\end{itemize}


\section{Related Work}

\subsection{Multimodal Table Understanding Datasets}
A variety of datasets have been introduced for multimodal table understanding (MTU), such as table recognition \cite{pubtabnet,fintabnet}, MMCoQA \cite{mmcoqa}, and MMTab \cite{MMTab}. 
Among them, MMTab is the most comprehensive, covering 24 tasks across four categories: Table Question Answering (TQA) \cite{WTQ,HiTab,tabmwp,zhu2021tatqa,aitqa,FeTaQA,tabmcq}, Table Fact Verification (TFV) \cite{TabFact,infotabs,pubhealthtab}, Table-to-Text generation (T2T) \cite{ToTTo,rotowire}, and Table Structure Understanding (TSU). 
Other works such as UniMMQA \cite{luo2023unifying}, MMTabQA \cite{mathur2024knowledge}, and MMTBench \cite{titiya2025mmtbench} further integrated visual features into table images for table-based QA and reasoning tasks. More recently, SynTab \cite{SynTab} scaled training resources to 1.2M samples by employing LLMs to synthesize question–answer pairs in short and detailed forms without executable verification, which may limit their consistency and correctness.

\subsection{(Multimodal) Large Language Models}
Large language models (LLMs) have recently been applied to table understanding via strategies such as prompt engineering \citep{table_cot,gpt4table}, instruction tuning \citep{tablellama,tablegpt_chatgpt,liu2023zero}, and tool integration \citep{lu2023chameleon,sheetcopilot,wangchain}, giving rise to models like TableLlama \citep{tablellama} and TableGPT \citep{tablegpt_chatgpt}. However, these approaches primarily process textual tables as inputs rather than table images.

Multimodal large language models (MLLMs) extend LLMs with visual capabilities. MLLMs such as Flamingo \cite{alayrac2022flamingo}, BLIP-2 \cite{blip2}, and LLaVA \cite{liu2023llava_1} have demonstrated strong performance on general multimodal benchmarks. Document-oriented models like Vary \cite{wei2023vary} and Monkey \cite{li2023monkey} scale to higher resolutions and complex layouts, while Qwen2-VL \cite{wang2024qwen2} and Qwen2.5-VL \citep{bai2025qwen2} are further trained on large-scale in-house datasets, achieving state-of-the-art results in document understanding. Nevertheless, MLLMs' performance remains limited on MMTab benchmarks. Finetuning MLLMs on MMTab and SynTab improves the performance, resulting in Table-LLaVA \cite{MMTab} and SynTab-LLaVA \cite{SynTab}.

Recent reasoning-augmented prompting and supervision have improved accuracy and interpretability in text-based domains, ranging from chain-of-thought \cite{wei2022chain} to code interpreter and program-of-thought \cite{mishra2022lila,chenprogram,gao2023pal}. Other research \cite{lu2023chameleon,suris2023vipergpt} employed LLMs with off-the-shelf modules to perform compositional reasoning and complex visual tasks.
For tabular reasoning, \citet{wangchain} introduced the Chain-of-Table framework that generates Python code sequences to perform complex operations on textual table data. For reasoning on table images, \citet{cheng2025vision} proposed an iterative chain-of-thought generation method to enhance table and document question answering. Building on this line of work, we are the first to introduce the code-driven reasoning framework and a tailored training pipeline to enable MLLMs to excel in MTU tasks across TQA, TFV, and TSU.

\begin{figure*}[t]
  \centering
  \includegraphics[width=0.98\linewidth]{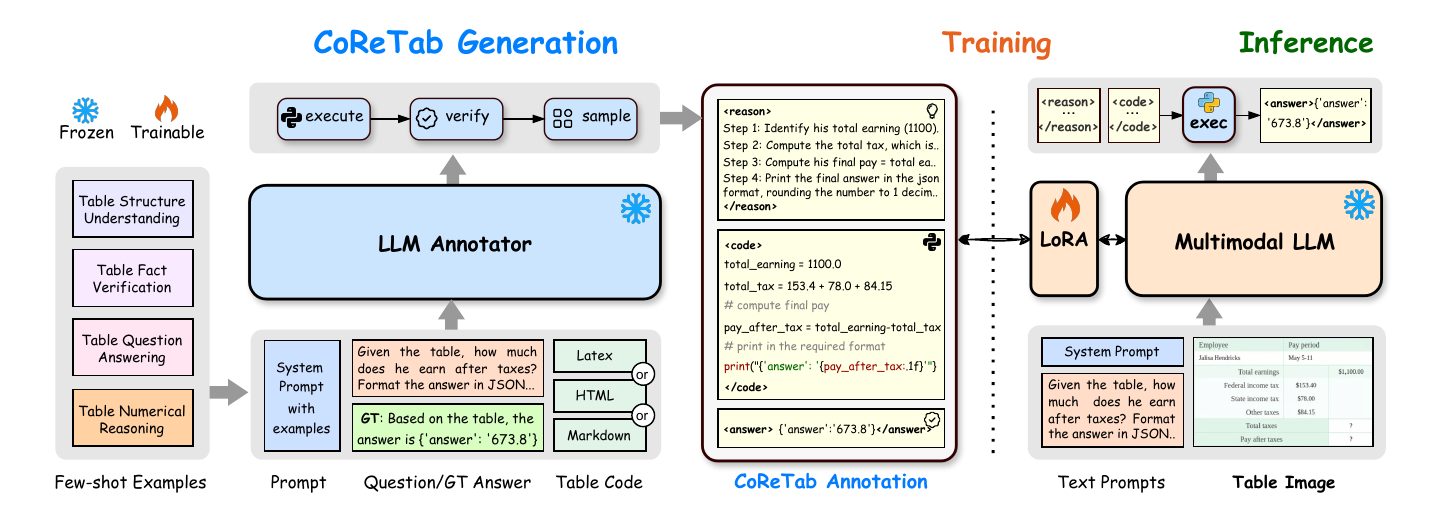}
  \caption{Overview of the \textbf{\CoReTab} framework. It uses an LLM to generate responses as new annotations with multi-step reasoning and Python code, which can be executed and verified against the ground truth with minimal human review. \CoReTab annotations provide a rich and interpretable supervision signal for training MLLMs.}
  \label{fig:coretab_framework}
  \end{figure*}

\section{\CoReTab Framework and Dataset}
\subsection{Data Collection}
\label{data_collection}
We construct \CoReTab from 11 publicly available datasets spanning three major task categories: Table Question Answering (TQA), Table Fact Verification (TFV), and Table Structure Understanding (TSU) (Table~\ref{tab:dataset_breakdown}). These tasks are associated with definitive answers and objective evaluation metrics such as Accuracy, which provide verifiable assessments of generated annotations. Following MMTab~\cite{MMTab}, we collect table codes (\LaTeX, HTML, or Markdown format) as part of the input to the LLM annotator, while the corresponding rendered images are part of ⟨table image, question, answer⟩ \CoReTab tuples. For comparability and guaranteed correctness of annotations, we reuse pairs of input questions and ground truth answers provided in MMTab, rather than SynTab, which may contain incorrect labels.

\subsection{Generation Framework}
We employ Qwen3 32B-3A \cite{yang2025qwen3}, a mixture-of-experts model with 3B active parameters out of 32B, as the LLM annotator. This LLM is both sufficiently powerful for our tasks and feasible to deploy on local GPUs, which ensures reproducibility and cost efficiency compared with proprietary closed-source models. Figure~\ref{fig:coretab_framework} shows the \CoReTab framework, in which the LLM annotator takes as input the tuple of 〈table code, question, its short GT answer〉and produces a structured \CoReTab annotation as a new GT answer. This new annotation provides better supervision signal for training MLLMs. 

\paragraph{Input and Prompting.} Each generation instance combines a textual table, a question, the ground-truth answer, and a structured system prompt as the input to the LMM annotator, as follows: 
\begin{tcolorbox}[
  title={\small \centering Input to the LLM annotator},
  colback=white,
  colframe=navy,
  colbacktitle=navy,
  coltitle=white,
  fonttitle=\bfseries,
  arc=1.5mm,      
  boxrule=1.0pt
]
\small
You are an expert in table analysis. Your task is to process a textual table to answer a question with a code-driven reasoning explanation. \\[0.3em]
You will be given few-shot examples to guide the expected reasoning and formatting style. \\[0.3em]
Your output must strictly follow this structure: \\[0.3em]
1. Reasoning — enclosed in \texttt{<reason>} tags. Provide clear, concise, and verifiable reasoning steps that logically lead to the final answer. \\[0.3em]
2. Python Code — enclosed in \texttt{<code>} tags. Write clean, executable Python code that computes the final answer based on the reasoning. The code must print the answer in the correct format. \\[0.3em]
3. Final Answer — enclosed in \texttt{<answer>} tags. This must contain the final answer or the printed output of the code snippet.\\[0.3em]
{\color{gray}\textbf{Tool Use}: \{\{ \texttt{Tool Use Descriptions} \}\} \# optional} \\[0.3em]
\textbf{Few-shot Examples}: \{\{ \texttt{Task examples} \}\} \\[0.3em]
\textbf{New Input}: \{\{ \texttt{table, question, GT answer} \}\}
\end{tcolorbox}
The system prompt is kept general across tasks, while task-specific few-shot examples are selected from a pool based on the question type and table format (HTML, \LaTeX, or Markdown). We further augment the prompt with descriptions of predefined functions (tool-use descriptions) following CodeAct \citep{CodeAct}, such as $\texttt{get\_cell\_content}$, when precise table operations are needed for table structure understanding (TSU) tasks; see {Appendix} for further details. 
\paragraph{Generation.} Conditioned on this input, the annotator produces a structured annotation consisting of: (1) natural-language reasoning enclosed in \reason tags, (2) executable Python code in \code tags, and (3) the final answer in \answer tags. Together, these components form a \CoReTab annotation that exposes intermediate reasoning steps while remaining automatically verifiable.

\paragraph{Post-processing.} Despite careful prompt design, LLMs may still produce plausible but incorrect reasoning. To address this, we apply further post-processing steps to filter out the hallucinated outputs. Specifically, we extract the Python code from each generated output, execute it, and compare its result against the ground-truth answer. Only outputs that pass this verification are retained. Additionally, we filter out responses with excessively long reasoning traces and apply down-sampling to maintain dataset balance. Overall, this post-processing procedure mitigates hallucinations and strives to ensure that the resulting annotations are verifiable, all with minimal human-in-the-loop intervention.

\subsection{Dataset Analysis}
\label{sec:dataset}

\begin{table}[t!]\small 
  \renewcommand{\arraystretch}{1.8}
  \setlength\tabcolsep{2pt}
  \centering
  \caption{Breakdown statistics of the \textbf{\CoReTab} dataset.}
  \label{tab:dataset_breakdown}
  
  \scalebox{0.73}{
  \begin{tabular}{c|c|c|c|c} 
  \hline

  {\textbf{Category}} & {\textbf{Task Name}} & {\textbf{Dataset}} & {\textbf{\# Samples}} & \textbf{Avg. Len} \\ 
  \hline
  \multirow{4}{*}{\begin{tabular}[c]{@{}c@{}}Table\\Question \\Answering\end{tabular}} & Flat TQA & WTQ (\citeyear{WTQ}) & 9.5K & 335 \\ 
  \cline{3-5}

   & {Hierarchical TQA} & HiTab (\citeyear{HiTab}) & 7.5K & 257 \\
  
  \cline{2-5}
  & \multirow{2}{*}{\begin{tabular}[c]{@{}c@{}}Tabular\\Numerical Reasoning\end{tabular}} & TABMWP (\citeyear{tabmwp}) & 30K & 287 \\
  \cline{3-5}
  &  & TAT-QA (\citeyear{zhu2021tatqa}) & 6K & 279 \\ 

  \cline{1-5}
  \multirow{2}{*}{\begin{tabular}[c]{@{}c@{}}Table Fact \\Verification \end{tabular}} & \multirow{2}{*}{TFV} & TabFact (\citeyear{TabFact}) & 22K & 347 \\
  \cline{3-5}
   &  & InfoTabs (\citeyear{infotabs}) & 15K & 347 \\
   \cline{1-5}
   \multirow{6}{*}{\begin{tabular}[c]{@{}c@{}}Table\\Structure\\ Understanding\\\end{tabular}} & Table Size Detection & TSD (\citeyear{MMTab}) & 5K & 632 \\ 
  \cline{2-5}
    & Table Cell Extraction & TCE (\citeyear{MMTab}) & 5K & 795 \\ 
  \cline{2-5}
    & Table Cell Locating & TCL (\citeyear{MMTab}) & 5K & 1427 \\ 
  \cline{2-5}
    & Merged Cell Detection & MCD (\citeyear{MMTab}) & 5K & 966 \\ 
  \cline{2-5}
    & Row\&Col Extraction & RCE (\citeyear{MMTab}) & 5K & 836 \\ 
  \cline{1-5}
   \multicolumn{3}{c|}{\textbf{Total}} & 115K & 529 \\ 
  \hline
  \end{tabular}
  }
  \end{table}
\label{dataset_analysis}
The final \CoReTab dataset comprises approximately 115K verified samples across 11 distinct multimodal table understanding tasks, as summarized in Table~\ref{tab:dataset_breakdown}. \CoReTab covers tables of varied structures, formats, and domains, providing a comprehensive and diverse resource for instruction-tuning multimodal LLMs on table understanding tasks. Compared to MMTab and SynTab, \CoReTab offers substantially richer annotations that explicitly integrate step-by-step reasoning and executable code. On average, \CoReTab responses contain 529 tokens—nearly $15\times$ longer than the short ground-truth answers in MMTab (32.6 tokens) and SynTab (34 tokens). This length reflects the multi-step reasoning and detailed derivation that the dataset encourages. See Appendix for more examples of \CoReTab annotations and other information.

\section{\CoReTab-Qwen2.5-VL}

  We adapt the pretrained Qwen2.5-VL \cite{bai2025qwen2} for multimodal table understanding with \emph{minimal architectural changes} and \emph{low additional computation}. 
  Unlike Table-LLaVA and SynTab-LLaVA, which pretrain models from scratch, we leverage Low-Rank Adaptation (LoRA) \citep{hu2022lora} to inject lightweight trainable parameters into the frozen backbone. LoRA adapters are inserted into the language model layers, enabling efficient finetuning without updating the base model. Each training stage employs \textit{its own set of LoRA adapters}, while adapters from previous stages remain frozen. As shown in Figure~\ref{fig:coretab_framework} (right), CoReTab-Qwen2.5-VL takes a 〈table image, question〉 pair as input and generates a structured \CoReTab response. A system prompt is prepended to the question. During inference, we prioritized the final answer produced by Python code execution; if no output was obtained, the answer enclosed in the \texttt{<answer>} tags was used instead.

\subsection{Training Pipeline}
\begin{figure}[t!]
  \centering
  \includegraphics[width=\columnwidth]{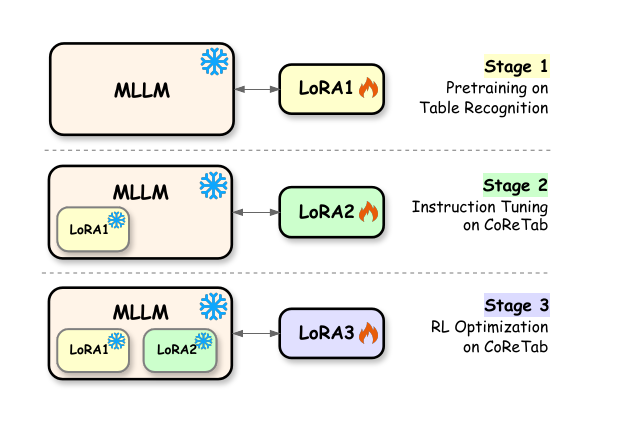}
  \caption{Our three-stage training pipeline: (1) pretraining on table recognition, (2) instruction tuning on \CoReTab, and (3) optionally RL optimization on \CoReTab. Separate LoRA adapters are updated in each stage.}
  \label{fig:pipeline}  
\end{figure}

 As shown in Figure~\ref{fig:pipeline}, our training pipeline comprises three stages: table recognition pretraining, instruction tuning on \CoReTab, and RL optimization on \CoReTab using GRPO. Each stage employs a new set of LoRA adapters, while only Stages 2 and 3 prepend a system prompt to the text input. We denote the final model merged with all three LoRA adapter sets as \CoReTab-Qwen2.5-VL.

  \paragraph{Table Pretraining.}  
  As shown in the top of Figure~\ref{fig:pipeline}, we first adapt Qwen2.5-VL on the MMTab-pre dataset of 150K table recognition samples \cite{MMTab}. The model learns to generate textual representations capturing both table structure and content, improving its ability to perceive table layouts and align multimodal features with tabular semantics. 
  
  \paragraph{Instruction Tuning.}  
  In the second stage, we perform supervised instruction tuning on 115K \CoReTab samples, covering diverse multi-task scenarios including question answering, reasoning, verification, and structure understanding. LoRA adapters from the pretraining stage are frozen, and a new set of adapters is optimized. We analyze the effect of sharing adapters in the ablation study.
  
  \paragraph{Reinforcement Learning.}  We further refine the model using GRPO \citep{grpo}, training it to maximize a reward function that combines accuracy and format alignment with \CoReTab preferences following \citep{guo2025deepseek}. LoRA adapters from prior stages remain frozen while a new set is updated. The influence of this stage is also tested in our ablation study.
  
\begin{table*}[t]\footnotesize
  \centering
  \renewcommand{\arraystretch}{1.15}
  \setlength\tabcolsep{2pt}
  \caption{Evaluation results on 7 MMTab benchmarks for TQA and TFV tasks. For all evaluation metrics, higher values indicate better performance. For Qwen2.5-VL-based models, the images are resized to a maximum of 1024 image tokens. `+Oracle' and `+OCR' denote whether the ground truth or OCR-extracted textual table representations are provided to LLMs, respectively.
  } 
  \label{tab:tqa_results}
  \scalebox{.88}{
  \begin{tabular}{lrccccccccc}
  \hline
  \multirow{3}{*}{\textbf{Method}} & \multirow{3}{*}{\textbf{LLM}} & \multicolumn{1}{c|}{\multirow{3}{*}{\textbf{Res.}}} & \multicolumn{5}{c|}{\textbf{Question Answering}} & \multicolumn{3}{c}{\textbf{Fact Verification}} \\
  \cline{4-11}
   &  & \multicolumn{1}{c|}{} & \multicolumn{1}{c|}{\textbf{TABMWP}} & \multicolumn{1}{c|}{\textbf{WTQ}} & \multicolumn{1}{c|}{\textbf{HiTab}} & \multicolumn{1}{c|}{\textbf{TAT-QA}} & \multicolumn{1}{c|}{\textbf{AIT-QA}} & \multicolumn{1}{c|}{\textbf{TabFact}} & \multicolumn{1}{c|}{\textbf{InfoTabs}} & \multicolumn{1}{c}{\textbf{PubHTab}} \\
   \cline{4-11}
   &  & \multicolumn{1}{c|}{} & \multicolumn{1}{c|}{Acc.} & \multicolumn{1}{c|}{Acc.} & \multicolumn{1}{c|}{Acc.} & \multicolumn{1}{c|}{Acc.} & \multicolumn{1}{c|}{Acc.} & \multicolumn{1}{c|}{Acc.} & \multicolumn{1}{c}{Acc.} \\
   \hline
  \multicolumn{11}{l}{\graycell\textbf{Open-source MLLM} (zero-shot)} \\
  LLaVA v1.5 7B & Vicuna-1.5 7B & 336 & 6.00 & 51.24 & 2.03 & 2.97 & - & 18.92 & 8.31 & - \\
  Vary-toy 2B & Qwen 1.8B & 1024 & 4.42 & 7.96 & 3.42 & 8.81 & 9.39 & 6.33 & 6.98 & - \\
  Monkey 7B & Qwen 7B & 896 & 13.26 & 19.07 & 6.41 & 12.31 & - & 22.56 & 22.11 & 18.89 \\
  Docowl1.5 7B & LLaMA2 7B & 448 & 11.41 & 26.80 & 11.10 & 12.44 & 46.18 & 27.67 & 28.74 & 28.42 \\
  IXcomposer2.5 7B & InternLM2 7B & 560 & 26.65 & 38.09 & 15.29 & 19.43 & 51.08 & 13.72 & 8.63 & 9.17 \\
  Ovis1.5 8B & LLaMA3 8B & 384 & 27.37 & 15.00 & 6.41 & 15.67 & 13.31 & 8.28 & 25.04 & 18.18 \\
  InternVL2 8B & InternLM2.5 7B & 448 & 16.51 & 25.09 & 7.68 & 13.99 & 23.28 & 28.32 & 36.63 & 37.23 \\
  Qwen2-VL 7B & Qwen2 7B & - & 26.04 & 37.38 & 25.19 & 22.67 & 67.12 & 16.29 & 39.67 & 32.60 \\
  TabPedia 7B & Vicuna-1.5 7B & 2560 & 12.27 & 20.37 & 1.22 & 9.71 & 17.22 & 28.84 & 9.20 & 21.01 \\
  Qwen2.5-VL 7B & Qwen2.5 7B & $1024^{*}$ & 54.69 & 55.23 & 38.09 &	56.88 & 50.59 & 71.37 & 70.21 & 70.15 \\[0.4em]
  \multicolumn{11}{l}{\graycell\textbf{LLMs} (zero-shot)}  \\
  LLaMA2+\textit{Oracle} & LLaMA 2 7B & - & 17.88 & 4.26 & 1.21 & 3.62 & - & 4.21 & 7.55 & - \\
  LLaMA2+\textit{OCR} & LLaMA 2 7B & - & 16.35 & 3.91 & 0.77 & 5.27 & - & 4.32 & 7.17 & -  \\
  TableLLaMA+\textit{Oracle} & LLaMA 2 7B & - & 12.98 & 31.63 & 64.71 & 2.84 & - & 82.55 & 2.85 & - \\
  TableLLaMA+\textit{OCR} & LLaMA 2 7B & - & 17.88 & 4.26 & 1.21 & 3.62 & - & 44.54 & 2.18 & -  \\
  \multicolumn{11}{l}{\graycell\textbf{Close-sourced MLLMs} (zero-shot)}  \\
  GPT-4V Low Res & Unkown & 512 & 60.00 & 22.50 & 9.50 & 19.50 & 19 & 45.50 & 58.50 & 59.50 \\
  GPT-4V Low Res & Unkown & 2000 & 60.50 & 48.00 & 27.50 & 32.50 & 62.5 & 45.50 & 65.60 & 67.00 \\ [0.4em] 
  \multicolumn{1}{l}{\graycell\textbf{Trained MLLMs}} & \multicolumn{1}{r}{\graycell \textbf{Dataset (\#Samples)}} & \multicolumn{9}{r}{\graycell} \\
  SynTab-LLaVA 7B & MMTab+SynTab (2.1M) & 1536 & 88.30 & 39.59 & 35.66 & 51.94 & 28.57 & 70.78 & 69.42 & 68.02 \\
  Table-LLaVA 7B & MMTab (232K) & 336 & 57.78 & 18.43 & 10.09 & 12.82 & 5.48 & 59.85 & 65.26 & 51.03 \\
  Table-LLaVA 13B & MMTab (232K) & 336 & 59.77 & 20.41 & 10.85 & 15.67 & 6.06 & 65.00 & 66.91 & 48.46 \\
  Table-Qwen2.5-VL 7B & MMTab (232K) & $1024^{*}$ & 91.85 & 55.51 & 66.68 & 65.45 & 76.61 & 77.02 & 69.52 & 76.96 \\
  \bluerow CoReTab Qwen2.5-VL 7B & \CoReTab (115K) & $1024^{*}$  & \textbf{97.14} & \textbf{60.93} & \textbf{67.45} & \textbf{82.60} & \textbf{81.57} & \textbf{84.28} & \textbf{76.04} & \textbf{80.21}\\
  \hline
  \end{tabular}
  }
\end{table*}
  
\begin{table*}[t]\footnotesize
  \centering
  \renewcommand{\arraystretch}{1.3}
  \setlength\tabcolsep{3pt}
  \caption{Evaluation results on 10 Table Structure Understanding benchmarks. For all evaluation metrics, higher values indicate better performance. We denote the OOD benchmarks with $\Phi$. For Qwen2.5-VL-based models, the images are resized to a maximum of 1024 image tokens.
  `+Oracle' and `+OCR' denote whether the ground truth or OCR-extracted textual table representations are provided to LLMs, respectively.
  } \label{tab:tsu_results}

  \scalebox{0.78}{
  \begin{tabular}{lrcccccccccccccc}
  \hline
  \multirow{3}{*}{\textbf{Method}} & \multirow{3}{*}{\textbf{LLM}} & \multicolumn{1}{c|}{\multirow{3}{*}{\textbf{Res.}}} & \multicolumn{2}{c|}{\textbf{TSD}} & \multicolumn{2}{c|}{\textbf{TSD$\Phi$}} & \multicolumn{1}{c|}{\textbf{TCE}} & \multicolumn{1}{c|}{\textbf{TCE$\Phi$}} & \multicolumn{1}{c|}{\textbf{TCL}} & \multicolumn{1}{c|}{\textbf{TCL$\Phi$}} & \multicolumn{1}{c|}{\textbf{MCD}} & \multicolumn{2}{c|}{\textbf{RCE}} & \multicolumn{2}{c}{\textbf{RCE$\Phi$}} \\ 
  \cline{4-16}
   &  & \multicolumn{1}{c|}{} & \multicolumn{1}{c|}{\begin{tabular}[c]{@{}c@{}}Row\\~Acc.\end{tabular}} & \multicolumn{1}{c|}{\begin{tabular}[c]{@{}c@{}}Col. \\Acc.\end{tabular}} & \multicolumn{1}{c|}{\begin{tabular}[c]{@{}c@{}}Row\\~Acc.\end{tabular}} & \multicolumn{1}{c|}{\begin{tabular}[c]{@{}c@{}}Col. \\Acc.\end{tabular}} &
    \multicolumn{1}{c|}{Acc.} & \multicolumn{1}{c|}{Acc.} &
    \multicolumn{1}{c|}{Acc.} & \multicolumn{1}{c|}{Acc.} &
    \multicolumn{1}{c|}{F1} & \multicolumn{1}{c|}{\begin{tabular}[c]{@{}c@{}}Row\\~F1\end{tabular}} & \multicolumn{1}{c|}{\begin{tabular}[c]{@{}c@{}}Col. \\F1\end{tabular}} & 
    \multicolumn{1}{c|}{\begin{tabular}[c]{@{}c@{}}Row\\~F1\end{tabular}} & \multicolumn{1}{c}{\begin{tabular}[c]{@{}c@{}}Col. \\F1\end{tabular}} \\ 

  \hline
  \multicolumn{16}{l}{\graycell\textbf{Open-source MLLMs} (zero-shot)} \\
  LLaVA v1.5 7B & Vicuna-1.5 7B & 336 & 0.80 & 2.50 & 2.40 & {-} & 0.22 & {-} & 0.62 & 0.93 & 1.26 & 1.66 & 4.13 & {-} & {-} \\ 
  Vary-toy 2B & Qwen 1.8B & 1024 & 1.30 & 2.20 & {-} & {-} & 1.96 & {-} & 0.73 & {-} & 0.52 & 2.01 & 2.38 & {-} & {-} \\ 
  Monkey 7B & Qwen 7B & 896 & 0.80 & 0.60 & {-} & {-} & 1.46 & 0.76 & 1.31 & {-} & 0.67 & 3.89 & 4.53 & 4.29 & {-} \\ 
  Docowl1.5 7B & LLaMA-2 7B & 448 & 0.20 & 1.50 & 0.40 & 0.80 & 1.00 & 1.80 & 0.00 & 0.00 & 0.00 & 2.40 & 4.60 & 0.60 & 1.30 \\ 
  IXcomposer-2.5 7B & InternLM2 7B & 560 & 1.60 & 8.00 & 2.40 & 12.40 & 3.66 & 3.80 & 1.25 & 3.13 & 2.30 & 0.17 & 0.09 & 0.00 & 0.00 \\ 
  Ovis1.5 8B & LLaMA-3 8B & 384 & 5.40 & 13.00 & 7.60 & 16.40 & 4.69 & 5.64 & 2.39 & 3.33 & 2.42 & 13.19 & 20.30 & 26.43 & 37.84 \\ 
  InternVL2 8B & InternLM2.5 7B & 448 & 7.70 & 34.10 & 12.40 & 40.80 & 11.59 & 12.36 & 4.86 & 9.45 & 0.79 & 1.46 & 9.86 & 1.11 & 6.95 \\ 
  Qwen2-VL 7B & Qwen2 7B & {-} & 4.70 & 15.30 & 5.20 & 20.40 & 9.10 & 8.35 & 3.86 & 6.13 & 1.03 & 17.14 & 23.07 & 21.52 & 30.39 \\ 
  TabPedia 7B & Vicuna-1.5 7B & 2560 & 2.80 & 10.20 & 4.80 & 12.40 & 0.16 & 0.11 & 0.00 & 0.00 & 0.00 & 1.53 & 3.73 & 1.62 & 1.21 \\ 
  Qwen2.5-VL 7B & Qwen2.5 7B & $1024^{*}$  & 27.90 & 59.60 & 44.00 & 55.20 & 31.12 & 42.62 & 14.33 & 19.57 & 0.26 & 40.71 & 61.77 & 38.90 & 59.96 \\ 
  \multicolumn{16}{l}{\graycell\textbf{LLMs} (zero-shot)} \\
  LLaMA2+\textit{Oracle} & LLaMA 2 7B & - & 1.70 & 3.60 & - & - & 0.62 & - & 0.17 & - & - & 9.36 & 18.03 & - & - \\
  LLaMA2+\textit{OCR} & LLaMA 2 7B & - & 1.30 & 3.40 & - & - & 0.35 & - & 0.15 & - & - & 8.35 & 10.45 & - & - \\
  TableLLaMA+\textit{Oracle} & LLaMA 2 7B & - & 5.30 & 4.40 & - & - & 9.35 & - & 0.82 & - & - & 4.34 & 18.03 & - & - \\
  TableLLaMA+\textit{OCR} & LLaMA 2 7B & - & 3.90 & 3.70 & - & - & 3.95 & - & 0.65 & - & - & 2.82 & 2.39 & - & - \\  
  \multicolumn{16}{l}{\graycell\textbf{Close-sourced MLLMs} (zero-shot)} \\
  GPT-4V-Low-res & Unknown & 512 & 6.00 & 24.00 & 8.00 & 15.00 & 3.57 & 10.29 & 14.41 & 17.73 & 2.12 & 30.32 & 56.86 & 27.69 & 50.36 \\ 
  GPT-4V-High-res & Unknown & 2000 & 12.50 & 46.00 & 19.00 & 38.00 & 9.75 & 14.36 & 23.38 & 27.91 & 3.50 & 26.44 & 43.17 & 48.52 & 57.14 \\ [0.4em] 
  \grayrow {\textbf{Trained MLLMs}} & \multicolumn{1}{r}{\textbf{Dataset (\# samples)}} & \multicolumn{14}{r}{} \\
  SynTab-LLaVA 7B & MMTab+SynTab (2.1M) & 1536 & 56.20 & 82.40 & 51.60 & 62.80 & 50.29 & 44.90 & 60.80 & 51.33 & 52.93 & 54.16 & 73.33 & 51.73 & 55.55 \\ 
  Table-LLaVA 7B & MMTab (232K) & 336 & 33.10 & 33.20 & 25.20 & 16.40 & 19.45 & 11.28 & 29.31 & 26.10 & 17.14 & 31.43 & 37.93 & 21.97 & 18.14 \\ 
  Table-LLaVA 13B & MMTab (232K) & 336 & 34.40 & 27.60 & 31.60 & 14.80 & 19.53 & 11.38 & 29.68 & 26.17 & 16.52 & 31.07 & 41.49 & 21.94 & 18.67 \\ 
  Table-Qwen2.5-VL 7B & MMTab (232K) & $1024^{*}$  & 52.90 & 78.30 & 52.80 & 74.80 & 58.67 & 64.64 & 63.11 & 67.64 & 39.65 & 65.13 & 77.71 & 72.01 & 83.97 \\ 
  \bluerow \CoReTab-Qwen2.5-VL 7B & \CoReTab (115K) & {$1024^*$} & \textbf{88.70} &	\textbf{87.60} & \textbf{85.60}	& \textbf{87.60} & \textbf{85.29} & \textbf{80.48}	& \textbf{87.41} &	\textbf{80.16}	& \textbf{82.92} &	\textbf{89.98} & \textbf{92.51} & \textbf{87.36} & \textbf{87.85} \\
  \hline
  \end{tabular}
  }
\end{table*}

\section{Experiments}
\label{experiments}

\subsection{Implementation Details}
\label{implementation_details}

All experiments use Qwen2.5-VL 7B as the base model. Three separate sets of LoRA adapters are applied sequentially across the three-stage training pipeline, allowing stage-specific adaptation while keeping previous adapters frozen. Each adapter is configured with a rank of 8 and alpha of 32 on all LLM linear layers. We optimize all trainable parameters using AdamW \citep{adamw} with a global batch size of 256 for 2 epochs per each stage. Input table images are resized to ensure the number of image tokens does not exceed 1024. Our training is performed on 8 H200 GPUs. See the Appendix for additional details.

\subsection{Experimental Setup}
\label{experimental_setup}

\paragraph{Baselines.} We compare \CoReTab against three categories of methods with results reported in prior work \citep{MMTab,SynTab} if available:

(1) \textbf{Open-source MLLMs.} These include LLaVA-1.5 \citep{improved_llava}, Vary-toy \citep{wei2024vary_toy}, Monkey \citep{li2023monkey}, mPLUG DocOwl-1.5 \citep{hu2024mplug_docowl}, Internlm Xcomposer-2.5 \citep{zhang2023internlmxcomposer}, Ovis-1.5 \citep{lu2024ovis}, InternVL2 8B \citep{chen2024expanding}, Qwen2-VL 7B \citep{wang2024qwen2}, and TabPedia \citep{zhao2024tabpedia}. We also include the zero-shot results of Qwen2.5-VL 7B \citep{bai2025qwen2}.  

(2) \textbf{Open-source LLMs.} We include LLaMA \cite{touvron2023llama2_model} and its counterpart TableLLaMA \cite{tablellama} with results reported in \cite{MMTab}.

(3) \textbf{Closed-source MLLM.} GPT-4V is included as a strong zero-shot baseline, with results reported on 100–200 randomly sampled examples per benchmark in \citet{MMTab}.  

(4) \textbf{Trained MLLMs.} We consider SynTab-LLaVA 7B (trained on SynTab \& MMTab), Table-LLaVA 7B/13B, and Qwen2.5-VL 7B trained on MMTab (denoted Table-Qwen2.5-VL 7B). We also report results from our \CoReTab-Qwen2.5-VL 7B trained on \CoReTab. All models were evaluated in single runs due to computational constraints.

\paragraph{Evaluation Benchmarks and Metrics.} 
We evaluate \CoReTab on 17 MMTab benchmarks \citep{MMTab} spanning three categories: Table Question Answering (TQA), Table Fact Verification (TFV), and Table Structure Understanding (TSU). 
Specifically, TQA includes WTQ, HiTab, TAT-QA, and TABMWP; TFV includes TabFact, InfoTabs, and PubHealthTab; and TSU includes TSD, Table Cell Extraction (TCE), Table Cell Locating (TCL), Merged Cell Detection (MCD), and  Row\&Column Extraction (RCE). 
For TSU, we additionally evaluate 4 out-of-distribution benchmarks (denoted as TSD$^\Phi$, TCE$^\Phi$, TCL$^\Phi$, and RCE$^\Phi$) whose tables do not appear in training.

We adopt task-specific metrics following~\citep{MMTab}.
For tasks with single-value ground truths (TQA, TFV, TCE, TCL), we report Accuracy after normalizing both predictions and references (e.g., lowercasing text, rounding numbers to the same precision). 
For TSD, we compute Accuracy separately for row and column counts. 
MCD and RCE are evaluated using cell-level F1 scores, with RCE reported independently for rows and columns. 
\subsection{Results and Analysis}
\label{results_and_analysis}

\paragraph{Table Question Answering Results.}
Table~\ref{tab:tqa_results} presents results on five question answering benchmarks: TABMWP, WTQ, HiTab, TAT-QA, and AIT-QA. Among open-source MLLMs, Qwen2.5-VL-7B achieves the highest accuracies, surpassing the closed-source GPT-4V on three benchmarks (WTQ, HiTab, and TAT-QA). However, it still lags behind models trained with task-specific datasets (i.e., SynTab, MMTab, and CoReTab).

Our \CoReTab-Qwen2.5-VL achieves the highest performance among all compared methods, scoring 97.14\% and 82.60\% on numerical reasoning tasks (TABMWP and TAT-QA), 60.93\% on WTQ, 67.45\% on HiTab, and 81.57\% on AIT-QA. It consistently improves results compared to Table-Qwen2.5-VL—specifically by over +14.96\% on TAT-QA—highlighting the benefit of code execution for precise numerical reasoning. Our model also surpasses SynTab-LLaVA by large margins, despite SynTab-LLaVA being trained on 2.1M samples, indicating that code-driven supervision can offset smaller data scales. Part of this gap may also stem from the stronger pretrained Qwen2.5-VL compared to LLaVA. 

Interestingly, Table-Qwen2.5-VL trained on MMTab also performs well on TABMWP, as MMTab inherits reasoning annotations from TABMWP, emphasizing the value of reasoning-augmented supervision for MLLMs.

\paragraph{Table Fact Verification Results.}
Table~\ref{tab:tqa_results} reports results on TabFact, InfoTabs, and PubHTab. Among zero-shot models, Qwen2.5-VL already surpasses the closed-source GPT-4V on all three benchmarks, underscoring the competitiveness of open-source MLLMs for factual reasoning tasks.

Finetuning with \CoReTab yields absolute gains of +12.91\% on TabFact, +5.83\% on InfoTabs, and +10.06\% on PubHTab. Compared with the MMTab-trained baseline (Table-Qwen2.5-VL), it achieves an average improvement of 5.65\% across three tasks. Our responses are also more transparent, derived from explicit reasoning steps and further verified with code execution, rather than only short answers such as ``confirmed'' or ``contradicted'' words.

\begin{figure*}[t]
  \centering
  \includegraphics[width=2.\columnwidth]{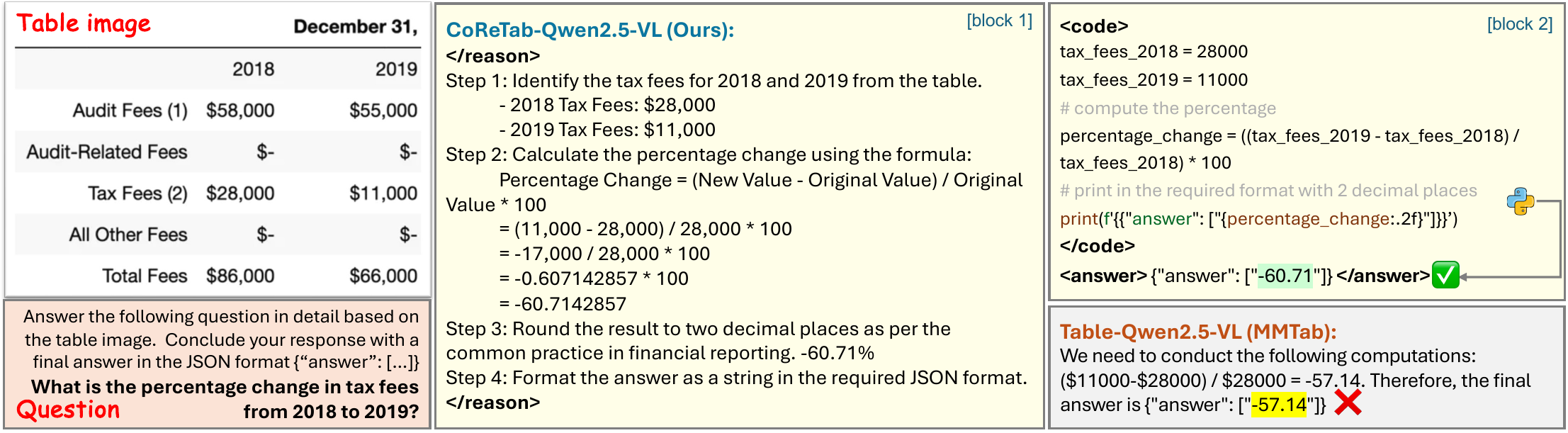}
  \caption{Comparison of responses generated by our \CoReTab-Qwen2.5-VL and the Table-Qwen2.5-VL baseline.
  }
  \label{fig:qualitative_results}  
\end{figure*}

\paragraph{Table Structure Understanding Results.}
Table~\ref{tab:tsu_results} reports results on 10 MMTab benchmarks for Table Structure Understanding, including 4 out-of-distribution benchmarks. Among open-source and close-sourced MLLMs in the zero-shot setting, Qwen2.5-VL 7B also attains the best performance such as reaching accuracies of 27.9/59.6 on TSD and 31.12 on TCE. This improves over earlier baselines such as InternVL2-8B and GPT-4v. 

Our \CoReTab-Qwen2.5-VL 7B substantially outperforms all baselines, achieving 88.70/87.60 on TSD, 85.29 on TCE, 87.41 on TCL, 82.92 on MCD, and 89.98/92.51 on RCE. It substantially outperforms Table-Qwen2.5-VL, achieving an average improvement of +25.56\%, highlighting the benefits of reasoning with executable code for accurately detecting table size, identifying merged cells, and localizing cell content.
Similar trends hold on out-of-distribution (OOD) tasks (i.e., TSD$^\Phi$, TCE$^\Phi$, TCL$^\Phi$, and RCE$^\Phi$).

\paragraph{LLM-only Performance.} As shown in Tables~\ref{tab:tqa_results} and~\ref{tab:tsu_results}, LLM-only methods (including LLaMA and TableLLaMA) substantially underperform multimodal LLMs, highlighting the necessity of MLLMs for effective table image understanding.

\paragraph{Qualitative Results.}
Figure~\ref{fig:qualitative_results} compares the responses generated by Qwen2.5-VL models trained on either \CoReTab or MMTab. The response from our \CoReTab-Qwen2.5-VL is accurate and interpretable, providing explicit reasoning traces and executable code for verifying intermediate steps and the final answer. 
In contrast, Table-Qwen2.5-VL produces a shorter, less informative response. Refer to our Appendix for more complex examples.

\paragraph{Prediction Analysis.}
We also analyzed randomly 100 model’s prediction errors per task. For TQA, 72 cases from cell extraction mainly due to small text in large tables or challenging table layouts; 22 cases from incorrect arithmetic logic; 3 cases related to translating reasoning to code.
For TFV tasks, 76 cases from cell extraction errors; the remainder from factual reasoning consistency.
For TSU tasks, 94 from cell extraction errors; only a small fraction from code-logic issues.

\subsection{Ablation study}

We conduct ablation experiments to assess the contributions of code-driven annotations and configurations to train Qwen2.5-VL with the \CoReTab dataset. Table~\ref{tab:ablation} reports average results across three benchmark categories—TQA, TFV, and TSU.

\paragraph{Effect of Code-driven Annotations.}
Removing the \code tags results in a large performance drop: $-4.19\%$ on TQA, $-3.26\%$ on TFV, and a striking $-17.96\%$ on TSU. This demonstrates that code-driven annotations are central to the success of \CoReTab, especially for TSU tasks that require precise localization and detection.

\paragraph{Full-parameter vs. LoRA Fine-tuning.}
Full-parameter fine-tuning reduces performance relative to LoRA training by $0.76\%$, $1.37\%$, and $1.58\%$ on TQA, TFV, and TSU, respectively. This suggests that full-parameter updates may disturb pretrained representations, while LoRA provides a more stable and efficient adaptation strategy. 
\paragraph{Shared vs. separate LoRAs.} We found that shared LoRA adapters across tasks (block (4)) achieve lower results (-1.2\% on average) compared to separate LoRA adapters for each stage, indicating that stage-specific LoRA training better preserves knowledge across the multi-stage pipeline.

\paragraph{Impact of Reinforcement Learning.}
Removing reinforcement learning leads to drops of $1.23\%$ and $0.84\%$ on TQA and TFV, but a slight gain (+$0.52\%$) on TSU. We interpret this as reinforcement learning being particularly effective for benchmarks that demand flexible multi-step reasoning (TQA, TFV), while its effect is limited—or even slightly negative—for TSU tasks, which can be solved with simpler reasoning and code.

\begin{table}[t]\footnotesize
  \centering
  \renewcommand{\arraystretch}{1.2}
  \setlength\tabcolsep{3pt}
  \scalebox{0.98}{
    \begin{tabular}{clccc}
      \hline
      \textbf{Setting} & \textbf{Configuration} & {\textbf{TQA}} & {\textbf{TFV}} & {\textbf{TSU}} \\
      \hline
      (1) &  \CoReTab-Qwen2.5-VL & {77.94} & {80.18} & {87.77} \\
      \hline
      \grayrow (2) &  w/o \code tags & 73.75 & 76.92 & 69.81 \\
        & $\Delta$ = (2) - (1) & -4.19 & -3.26 & -17.96 \\
      \hline
      \grayrow (3) &  with full-parameter training & 77.18 & 78.81 & 87.19 \\
          & $\Delta$ = (3) - (1) & -0.76	& -1.37	& -0.58 \\
      \hline
      \grayrow (4) &  with shared LoRA adapters & 76.84 & 79.25 & 86.14 \\
          & $\Delta$ = (4) - (1) & -1.10	& -0.93	& -1.63 \\
      \hline
      \grayrow (5) & w/o RL optimization & 76.71	& 79.34 & 88.29 \\
          & $\Delta$ = (5) - (1) & -1.23 & -0.84 & +0.52 \\
      \hline\hline
      \grayrow (6.1) & InternVL3 8B + MMTab & 67.32 & 71.28	& 61.32 \\
      (6.2) & InternVL3 8B + CoReTab & 72.58	& 77.14	& 74.64 \\
      & $\Delta$ = (6.1) - (6.2) & 5.26	& 5.86	& 13.32 \\
      \hline
      \end{tabular}
  }
  \caption{
  Ablation experiment results, reporting average performance across three benchmark categories. 
  }\label{tab:ablation}
\end{table}

\paragraph{Impact on other MLLMs.} We evaluate our framework on InternVL3-8B (Block 6 in Table~\ref{tab:ablation}). The model exhibits the same performance trend as Qwen2.5-VL, showing consistent improvements when trained on CoReTab.

\section{Conclusion}
We presented \textbf{\CoReTab}, a code-driven reasoning framework for multimodal table understanding that combines natural-language multi-step reasoning with executable code. By generating scalable, interpretable, and verifiable annotations, \CoReTab provides richer supervision to train MLLMs than existing datasets. We have shown that models trained on the \CoReTab achieve consistent improvements across 17 MMTab benchmarks, particularly in multi-step reasoning and table structure understanding, while producing transparent, verifiable, and human-aligned responses. Overall, our \CoReTab framework can be effectively applied to multimodal table understanding, enhancing both accuracy and interpretability.

\section{Acknowledgement}
This work was partly supported by JSPS KAKENHI Grant Number 23H00482. This work was written by the authors, with minor phrasing and grammatical polishing provided by ChatGPT. All interpretations and remaining errors are solely the authors’ responsibility.

\section{Limitations}
\label{sec:limitations}
While our \CoReTab framework provides promising supervision that improves both accuracy and interpretability for MTU, several limitations remain discussed.

First, some reasoning annotations inside \texttt{<reason>} tags of verified samples may still contain errors, as they are automatically synthesized by the LLM annotator and verified through direct output comparison without exhaustive human checks.
Generating long responses with code-driven reasoning increases inference time. Also, in certain cases, particularly in information retrieval and fact verification tasks, \texttt{<code>} tags may simply print outputs from \texttt{<reason>}, reducing the effectiveness of code verification.

Second, our dataset is curated primarily from academic sources in English (e.g., \cite{tabmwp, TabFact}) with clean table codes in HTML, \LaTeX, and Markdown. In industrial settings, table images may exhibit rotation, occlusion, deformation, or perspective distortions, making accurate table code extraction dependent on robust off-the-shelf table recognition models.

Lastly, \CoReTab focuses on tasks with deterministic, verifiable answers (e.g., associated with Accuracy and F1 metrics). Table text generation tasks such as summarization or description remain unexplored; future work could adopt semantic evaluation methods, such as LLM-as-a-judge \cite{zheng2023judging}, to extend verification beyond automatic metrics like BLEU.

\section{Ethical Considerations}
\label{sec:ethical_considerations}

The proposed \CoReTab dataset is built upon MMTab annotations and other publicly available academic datasets such as WTQ and TabFact, which are distributed under permissive licenses (e.g., MIT, CC-BY-SA-4.0). We do not introduce or collect any new personal information or offensive content beyond what may already exist in the underlying datasets. Consequently, any biases present in the annotations or model outputs reflect those in the source datasets and models.

Our framework is intended solely for research purposes, enabling the development of more robust multimodal table reasoning models, which may also be extended to related domains. For the generated code in \CoReTab, we follow the safeguards in \cite{CodeAct}, allowing compilation and execution only for data manipulation and basic numerical or logical operations. Potentially harmful actions (e.g., accessing modules such as \texttt{os} or \texttt{sys} to delete or modify files) are explicitly prohibited.

\bibliography{custom}

\clearpage
\appendix

\section{Details about CoReTab}
\label{more_details_of_dataset_construction}

The \CoReTab dataset and trained models will be publicly released to facilitate reproducibility and further research on multimodal table understanding. All latest resources will be made available upon publication.

\subsection{Additional Generation Details}
We use the vLLM engine \cite{kwon2023efficient} to accelerate inference. The sampling temperature is set to 0.1 for TSU tasks to ensure code correctness, and 0.3 for other tasks to encourage response diversity. As mentioned in the main paper, we augment the input to the LLM annotator with \textit{Tool Use} text, which describes the predefined functions available for TSU tasks. See Figure~\ref{fig:tool_use} for details.

For TSU tasks, we sort the responses by total token length and select the first 5,000 samples to avoid lengthy annotations. For TQA and TFV tasks, we further filter out samples with excessively long text inside \texttt{reason} or \texttt{code} tags, removing any response where the content within either tag exceeds 1,024 tokens.

\subsection{Additional Dataset Statistics}
\begin{figure*}[ht]\centering
  \includegraphics[width=2.1\columnwidth]{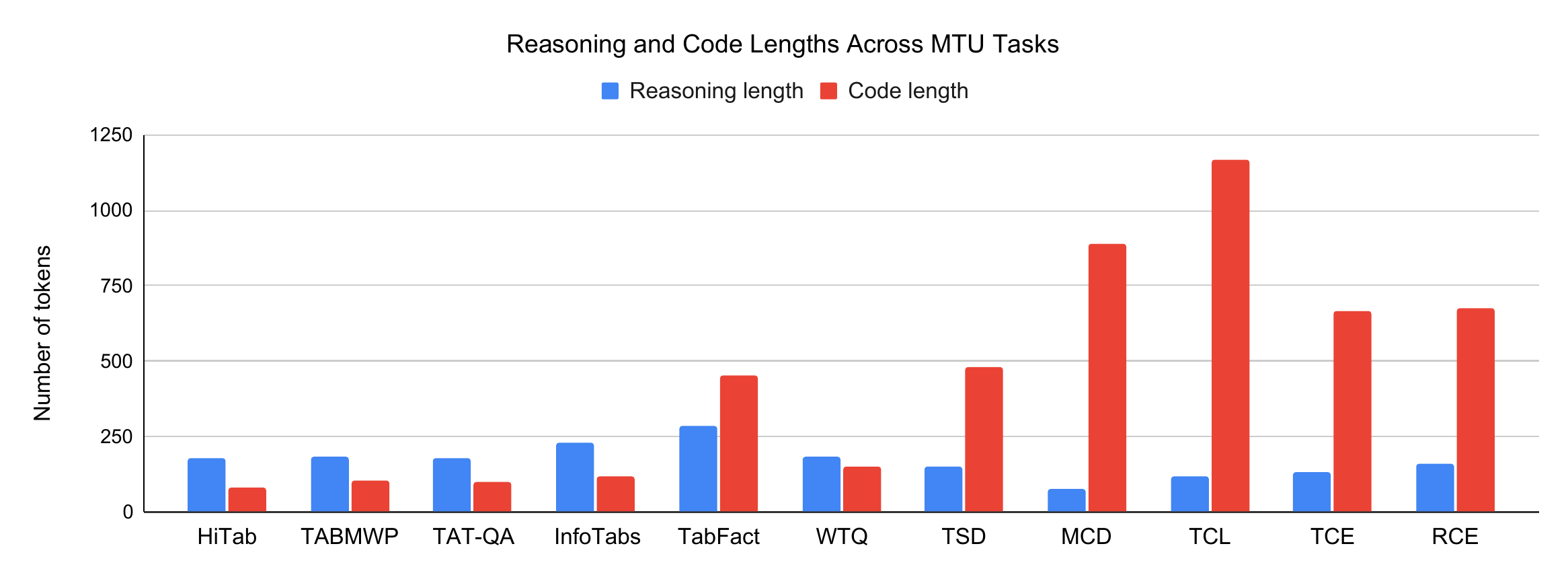}
    \caption{Length of tokens inside \reason and \code tags over 11 annotated tasks.}
  \label{fig:length_dist}
\end{figure*}

Figure~\ref{fig:length_dist} shows the distribution of reasoning lengths and Python code lengths across the 11 annotated tasks in the CoReTab dataset. We observe that for most TQA and TFV tasks, the average Python code is shorter than the corresponding multi-step reasoning explanation, whereas TSU tasks typically produce longer Python code that includes table textual representation for precise detection and localization.

\begin{figure*}[ht]\centering
  \includegraphics[width=2.1\columnwidth]{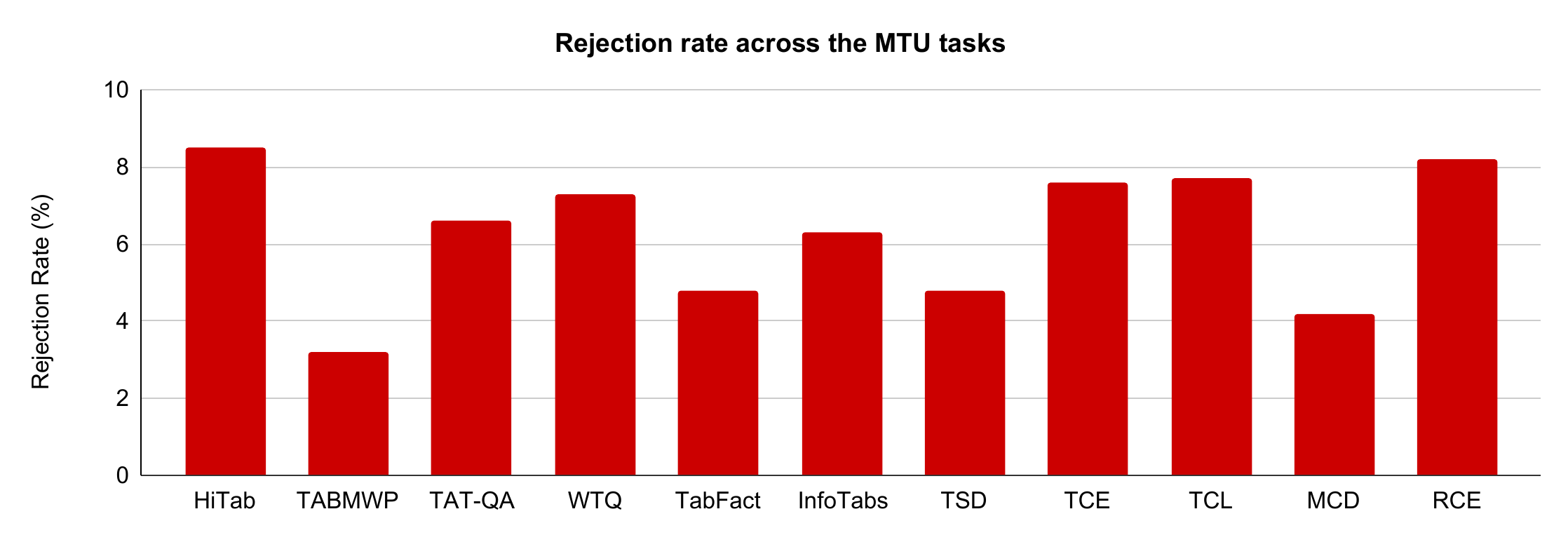}
    \caption{Rejection rates of generated annotations by the LLM over multimodal table understanding tasks.}
  \label{tab:rejection-rates}
\end{figure*}

Figure~\ref{tab:rejection-rates} shows the rejection rates of the responses generated by the LLM annotator. To accept or reject a response as a \CoReTab annotation, we extract its generated Python code and execute it for verification against the ground-truth answers. Here, the rejection rate refers to the proportion of responses whose Python code produces values inconsistent with the corresponding ground-truth answers.

\begin{figure}[t]
    \centering
\begin{tcolorbox}[
  title={\footnotesize \centering Descriptions of pre-defined functions},
  colback=white,
  colframe=navy,
  colbacktitle=navy,
  coltitle=white,
  fonttitle=\bfseries,
  arc=1.5mm,      
  boxrule=1.0pt
]\footnotesize
Use the following tools to assist in reaching the final answer:
\begin{itemize}[leftmargin=1.0em]
    \item \texttt{get\_table\_2d}: Get the 2D grid of the HTML table. Args: \texttt{html\_table}: \texttt{str}. Return: \texttt{list}.
    \item \texttt{get\_cell\_location}: Get the location of a cell in the table. Args: \texttt{table\_2d}: \texttt{list}, \texttt{row\_id}: \texttt{int}, \texttt{col\_id}: \texttt{int}. Return: \texttt{str}.
    \item \texttt{get\_table\_size}: Get the size of the table. Args: \texttt{table\_2d}: \texttt{list}. Return: \texttt{str}.
    \item \texttt{get\_cell\_value}: Get the value of a cell in the table. Args: \texttt{table\_2d}: \texttt{list}, \texttt{row\_id}: \texttt{int}, \texttt{col\_id}: \texttt{int}. Return: \texttt{str}.
    \item \texttt{get\_row\_values}: Get the values of a row in the table. Args: \texttt{table\_2d}: \texttt{list}, \texttt{row\_id}: \texttt{int}. Return: \texttt{str}.
    \item \texttt{get\_col\_values}: Get the values of a column in the table. Args: \texttt{table\_2d}: \texttt{list}, \texttt{col\_id}: \texttt{int}. Return: \texttt{str}.
    \item \texttt{get\_merged\_cell\_locations}: Get the locations of merged cells in the table. Args: \texttt{html\_table}: \texttt{str}. Return: \texttt{str}.
\end{itemize}
\end{tcolorbox}
    
    \caption{Descriptions of pre-defined functions used for TSU tasks to the (M)LLM.}
    \label{fig:tool_use}
\end{figure}



\subsection{Dataset Examples}
\label{more_dataset_examples}
\begin{figure*}[ht]\centering
  \includegraphics[width=1.5\columnwidth]{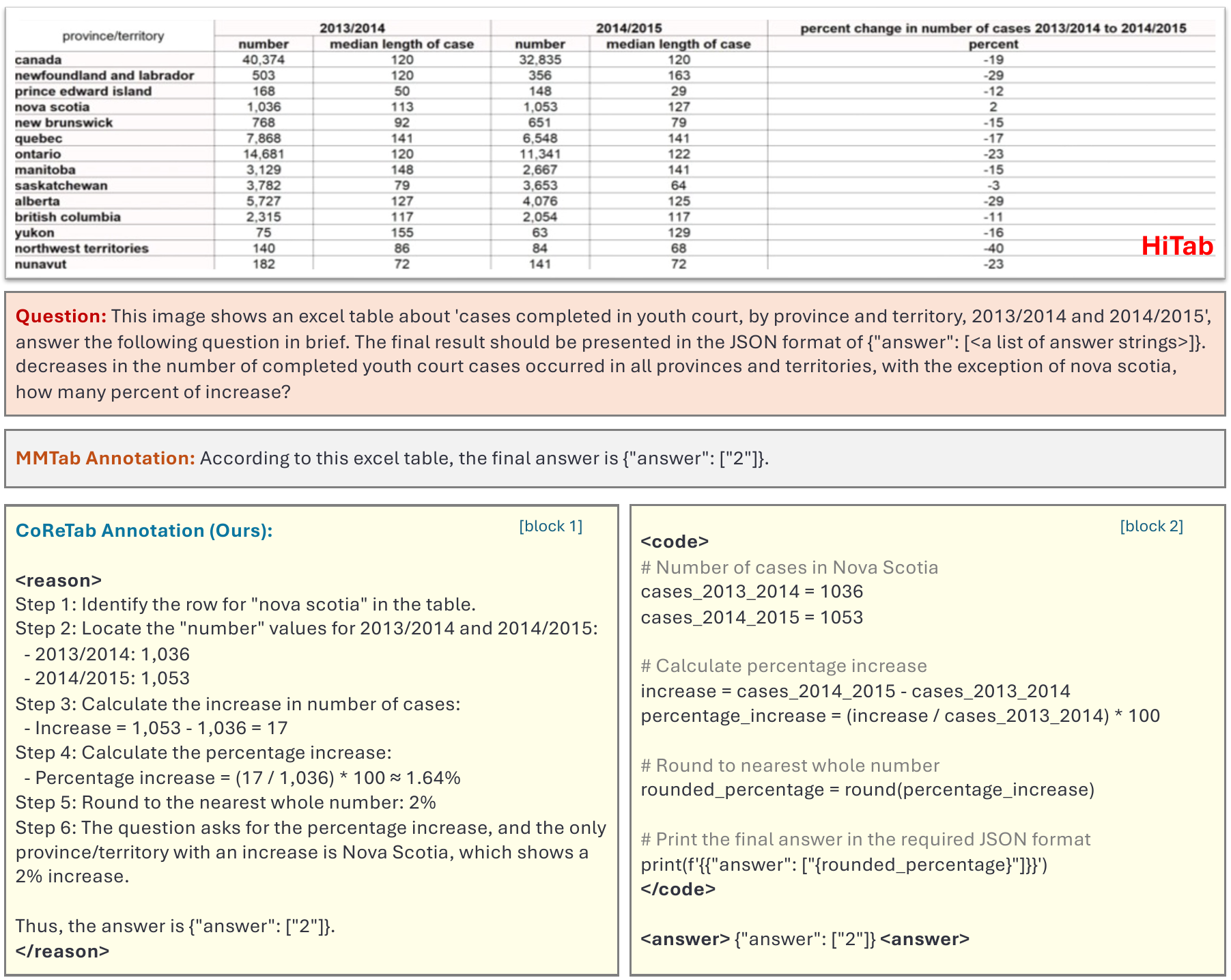}
    \caption{MMTab and \CoReTab annotations for a sample from HiTab.}
  \label{fig:extra_example1}
\end{figure*}

\begin{figure*}[ht]\centering
  \includegraphics[width=1.5\columnwidth]{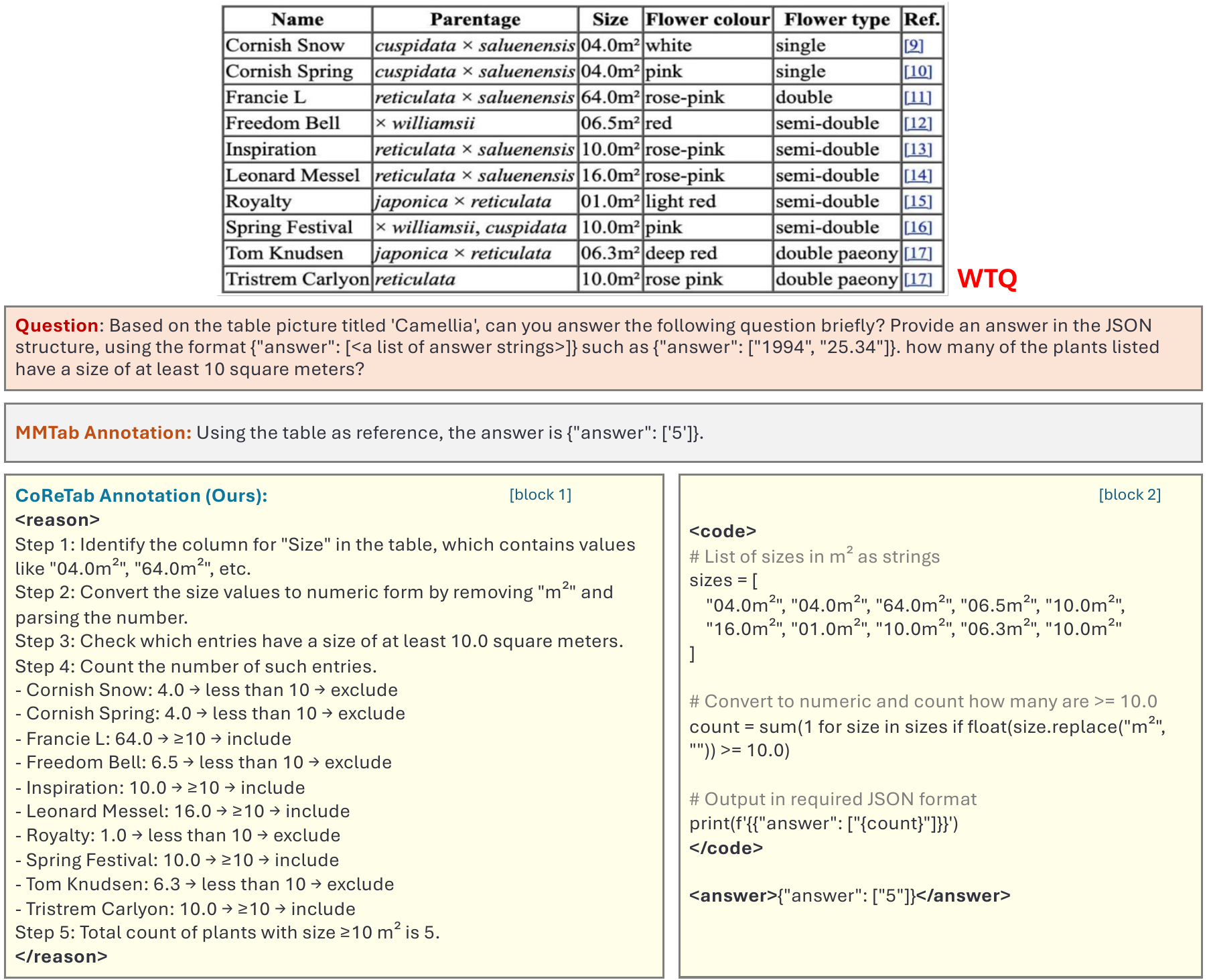}
    \caption{MMTab and \CoReTab annotations for a sample from WTQ.}
  \label{fig:extra_example2}
\end{figure*}

\begin{figure*}[ht]\centering
  \includegraphics[width=1.5\columnwidth]{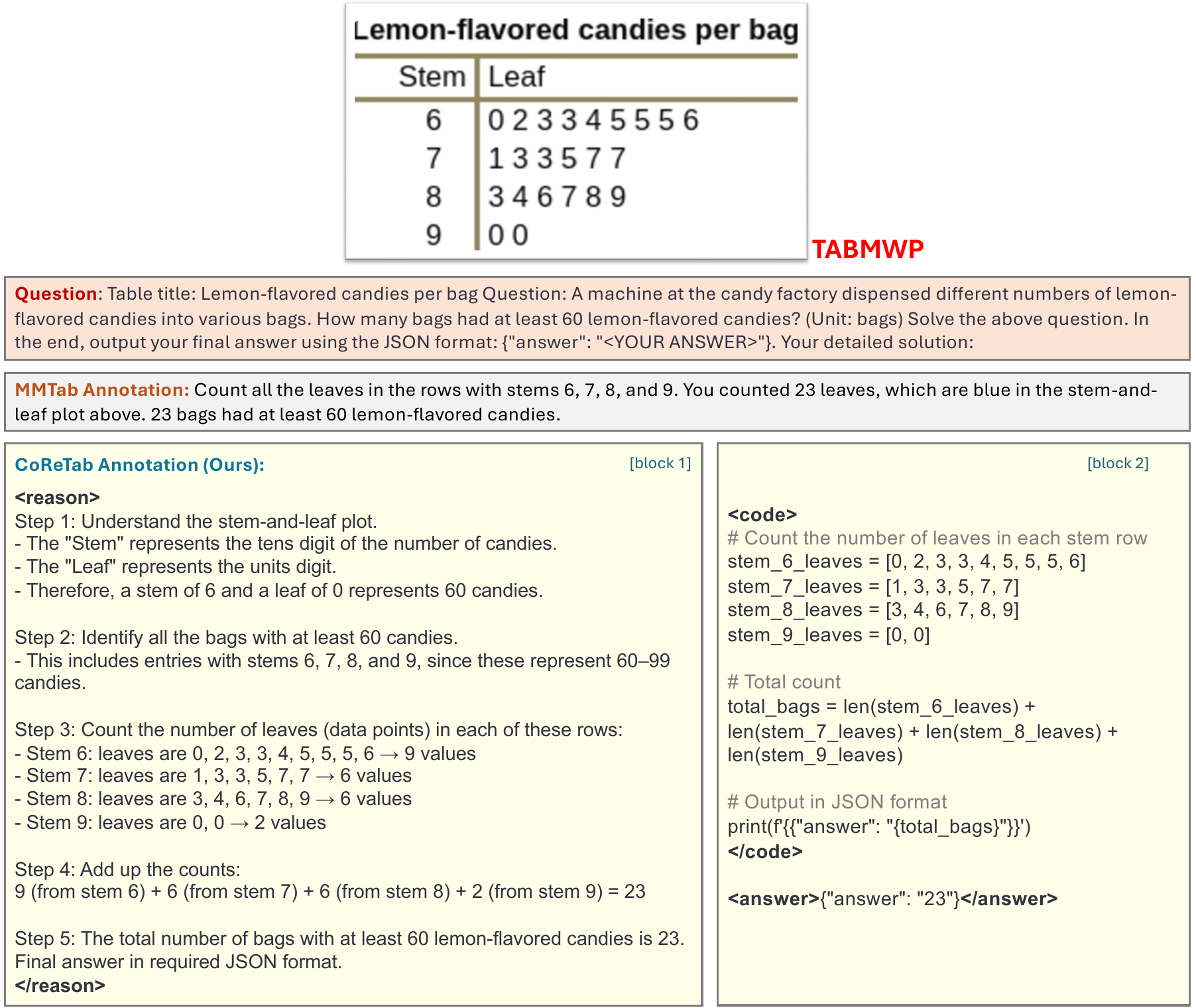}
    \caption{MMTab and \CoReTab annotations for a sample from TABMWP.}
  \label{fig:extra_example3}
\end{figure*}

\begin{figure*}[ht]\centering
  \includegraphics[width=1.5\columnwidth]{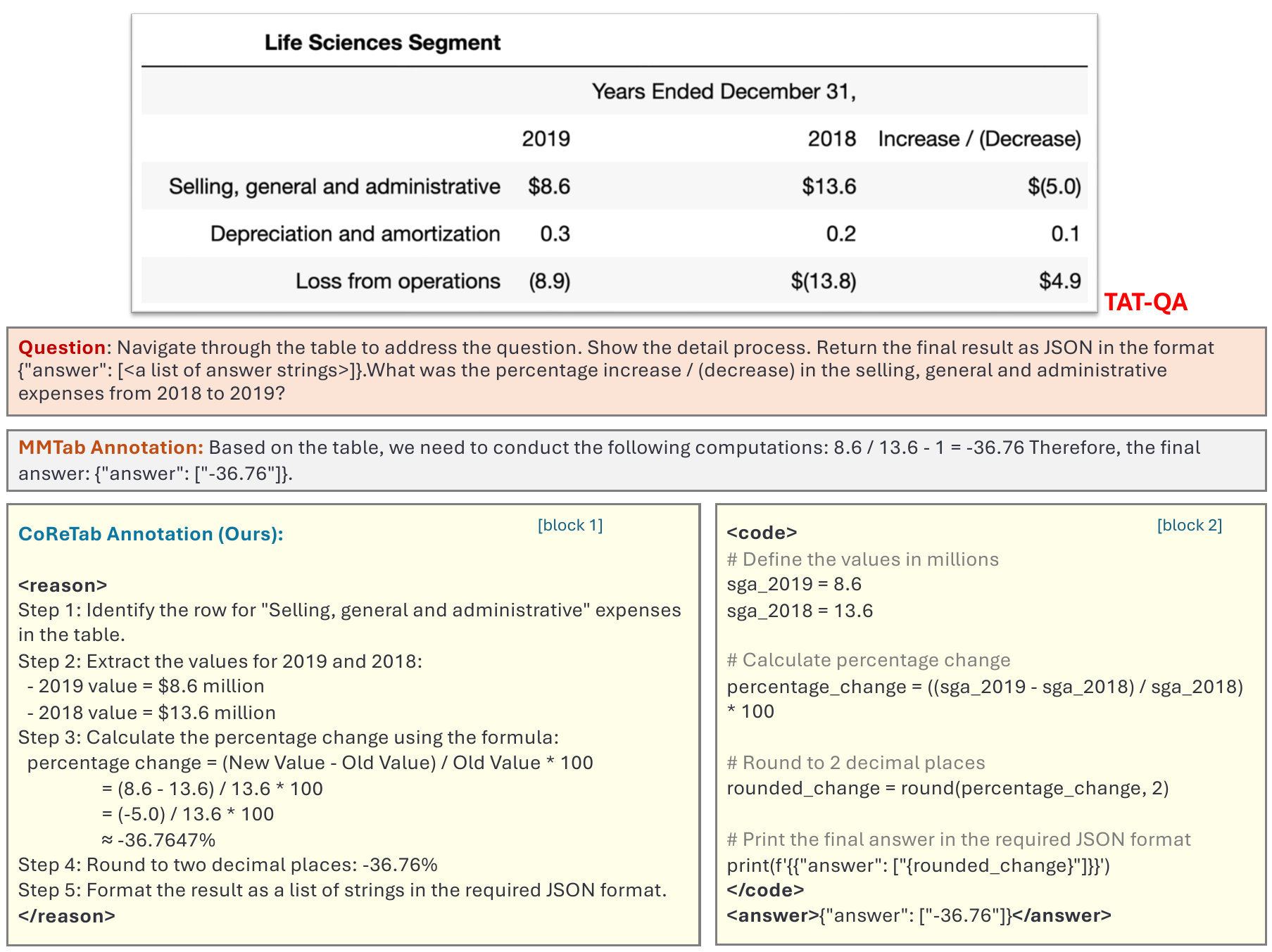}
    \caption{MMTab and \CoReTab annotations for a sample from TAT-QA.}
  \label{fig:extra_example4}
\end{figure*}

\begin{figure*}[ht]\centering
  \includegraphics[width=1.5\columnwidth]{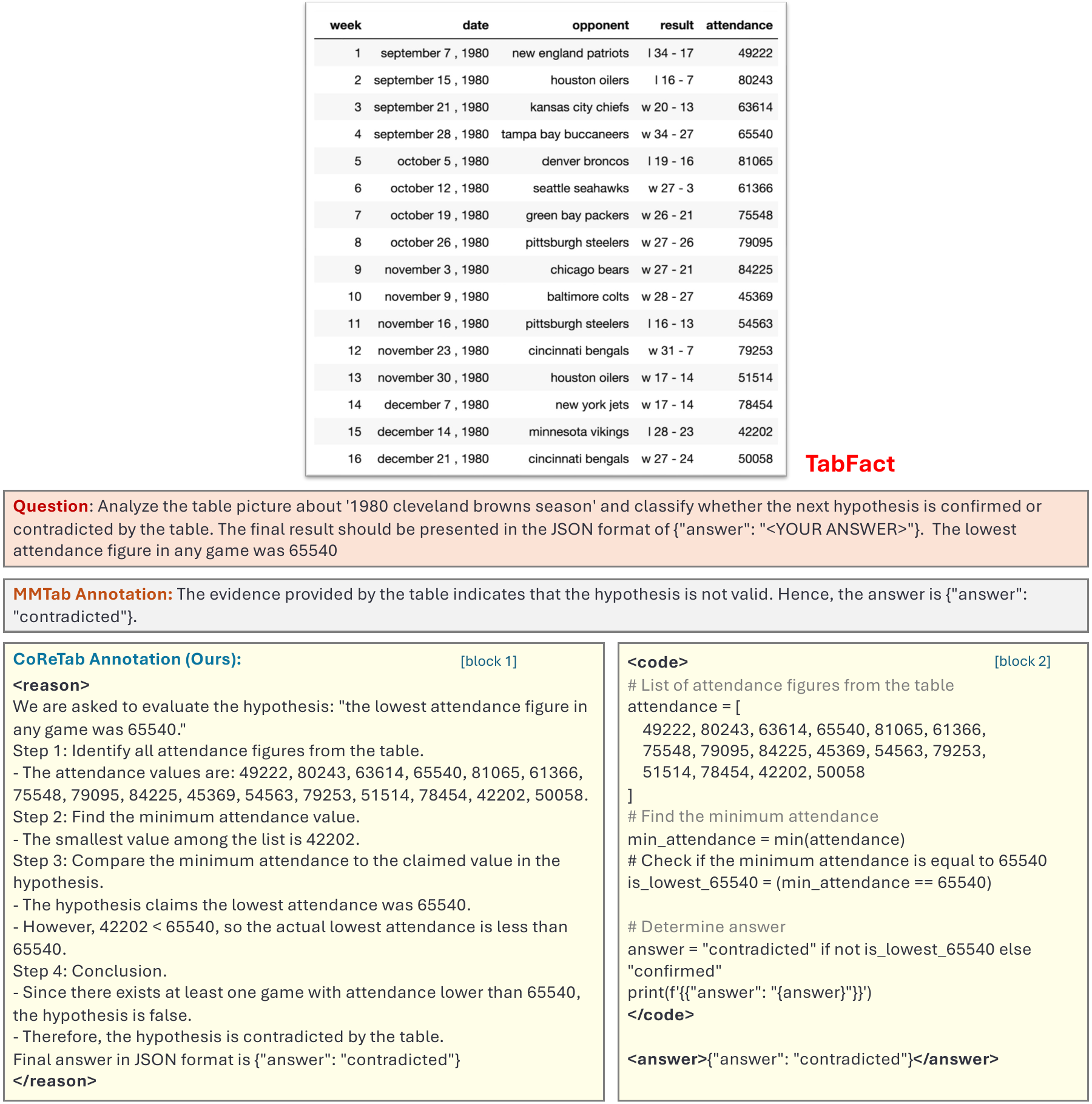}
    \caption{MMTab and \CoReTab annotations for a sample from TabFact.}
  \label{fig:extra_example5}
\end{figure*}

\begin{figure*}[ht]\centering
  \includegraphics[width=1.5\columnwidth]{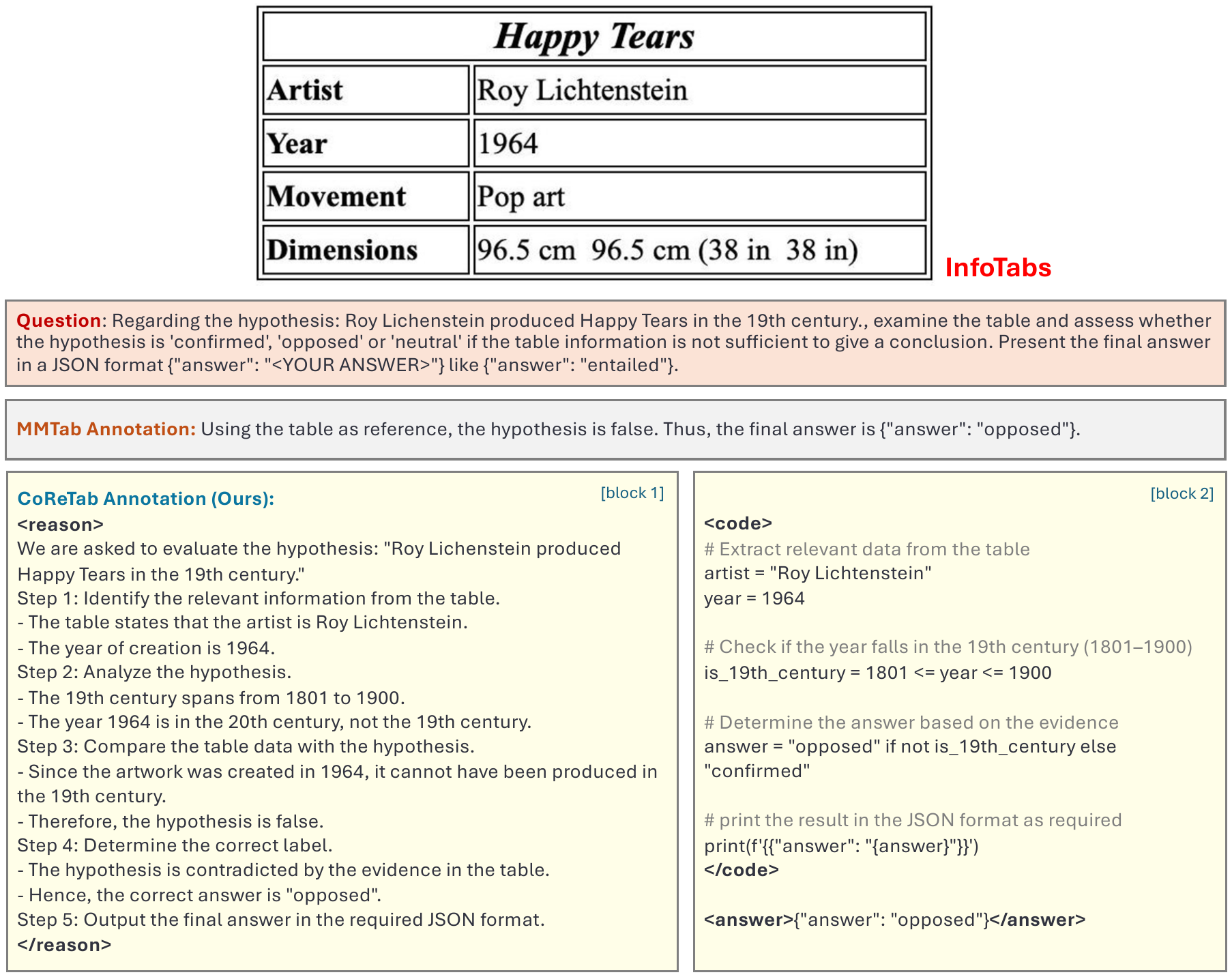}
    \caption{MMTab and \CoReTab annotations for a sample from InfoTabs.}
  \label{fig:extra_example6}
\end{figure*}

\begin{figure*}[ht]\centering
  \includegraphics[width=1.5\columnwidth]{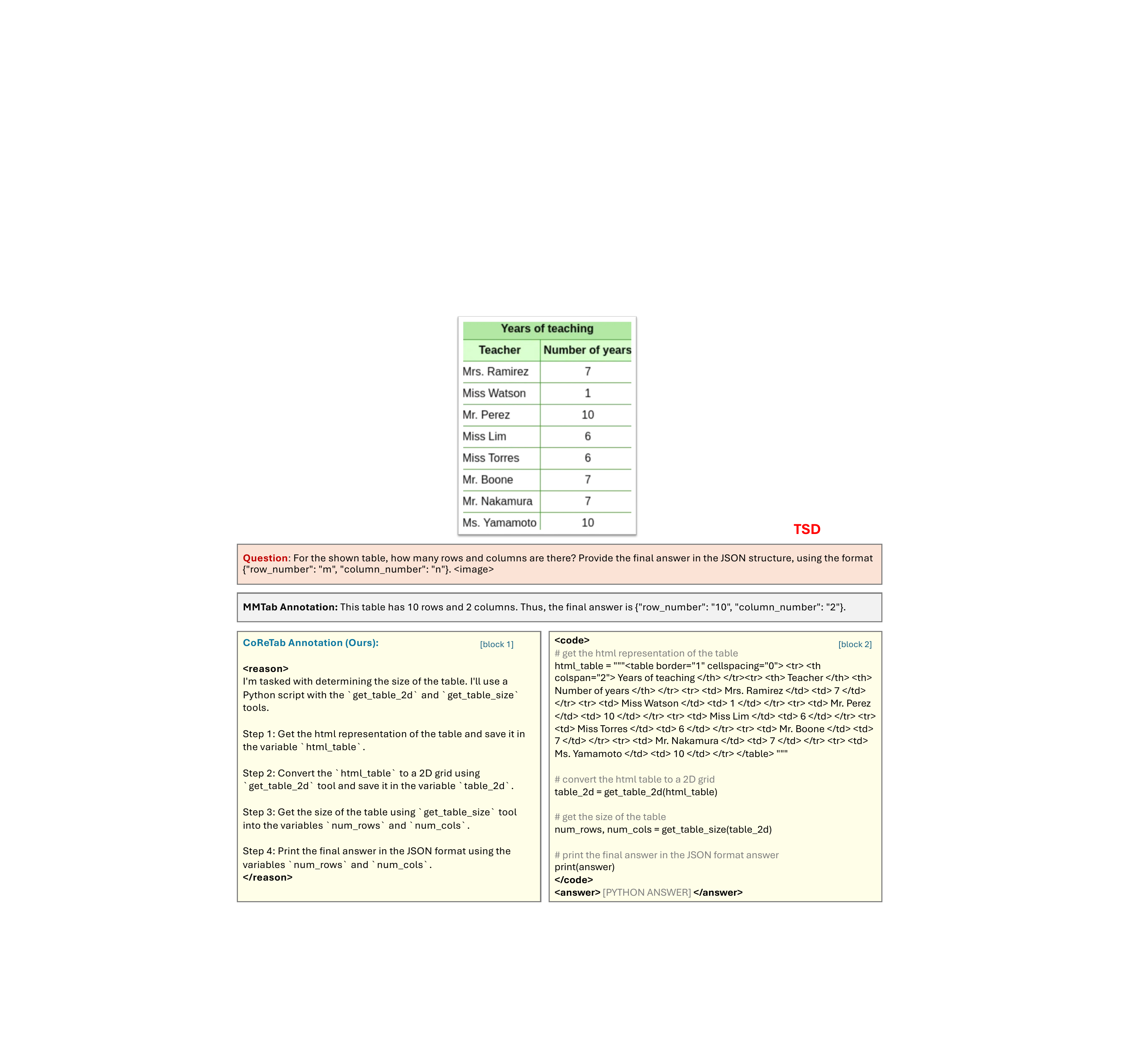}
    \caption{MMTab and \CoReTab annotations for a sample from TSD.}
  \label{fig:extra_example7}
\end{figure*}

\begin{figure*}[ht]\centering
  \includegraphics[width=1.5\columnwidth]{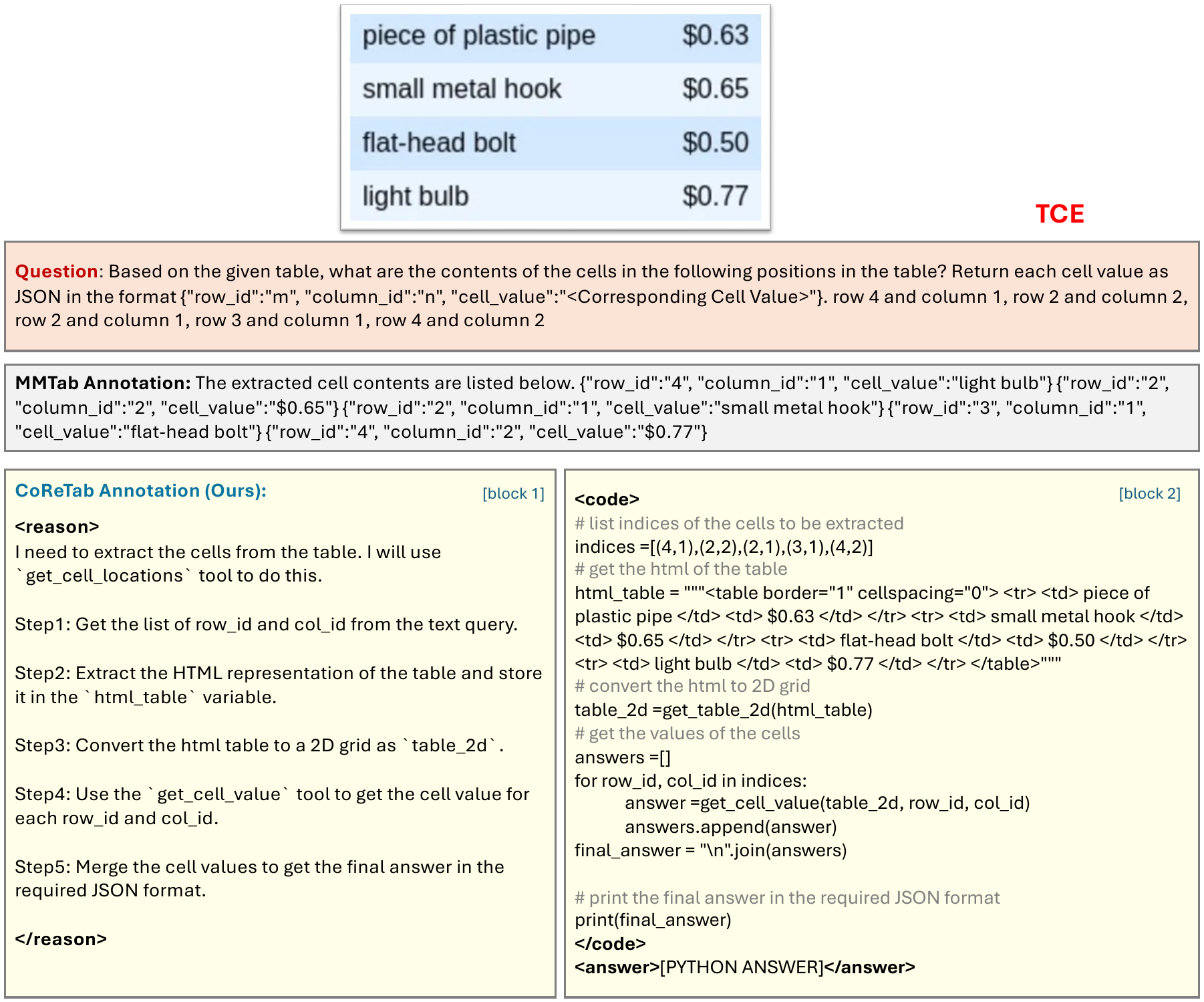}
    \caption{MMTab and \CoReTab annotations for a sample from TCE.}
  \label{fig:extra_example8}
\end{figure*}

\begin{figure*}[ht]\centering
  \includegraphics[width=1.5\columnwidth]{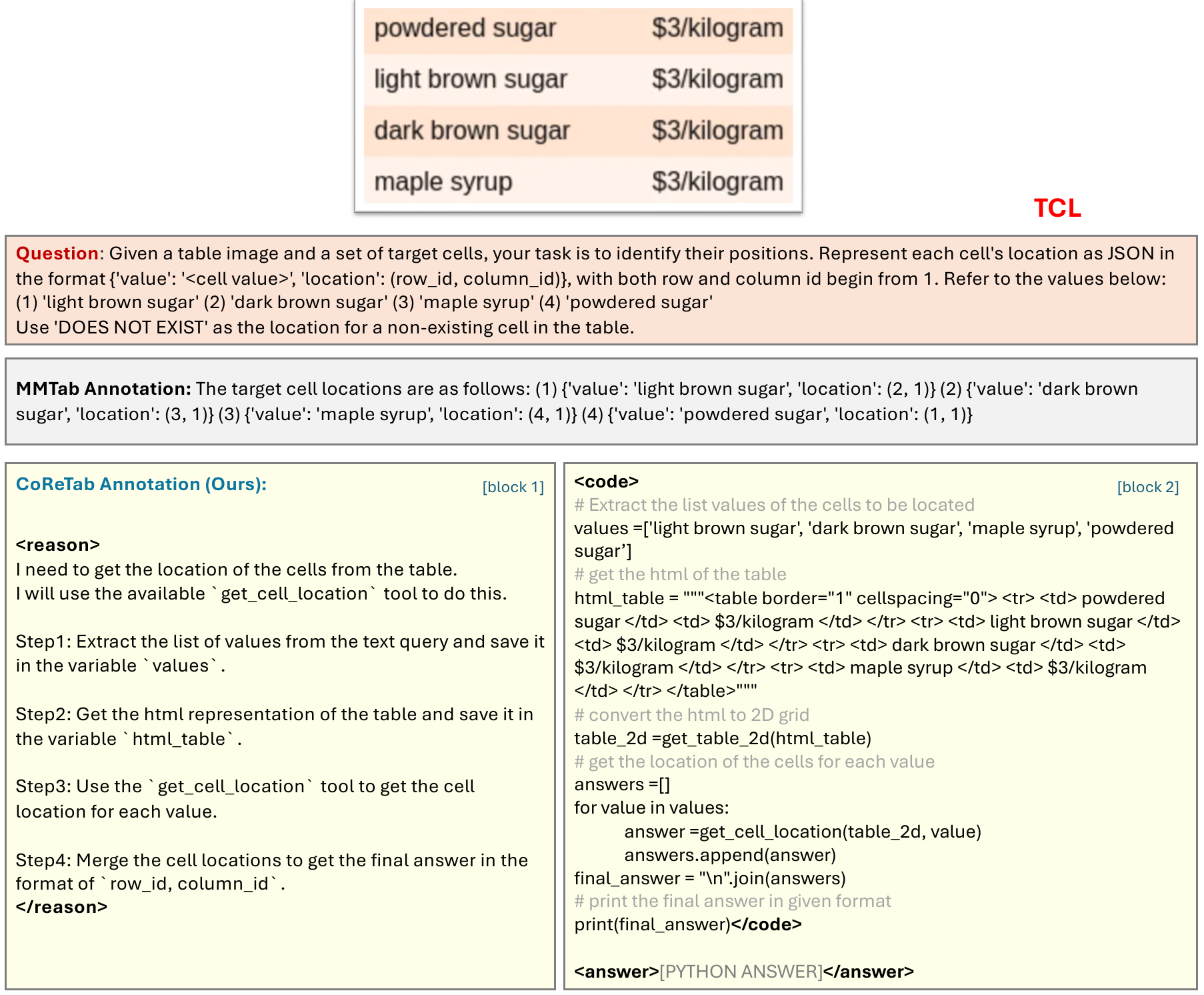}
    \caption{MMTab and \CoReTab annotations for a sample from TCL.}
  \label{fig:extra_example9}
\end{figure*}

\begin{figure*}[ht]\centering
  \includegraphics[width=1.5\columnwidth]{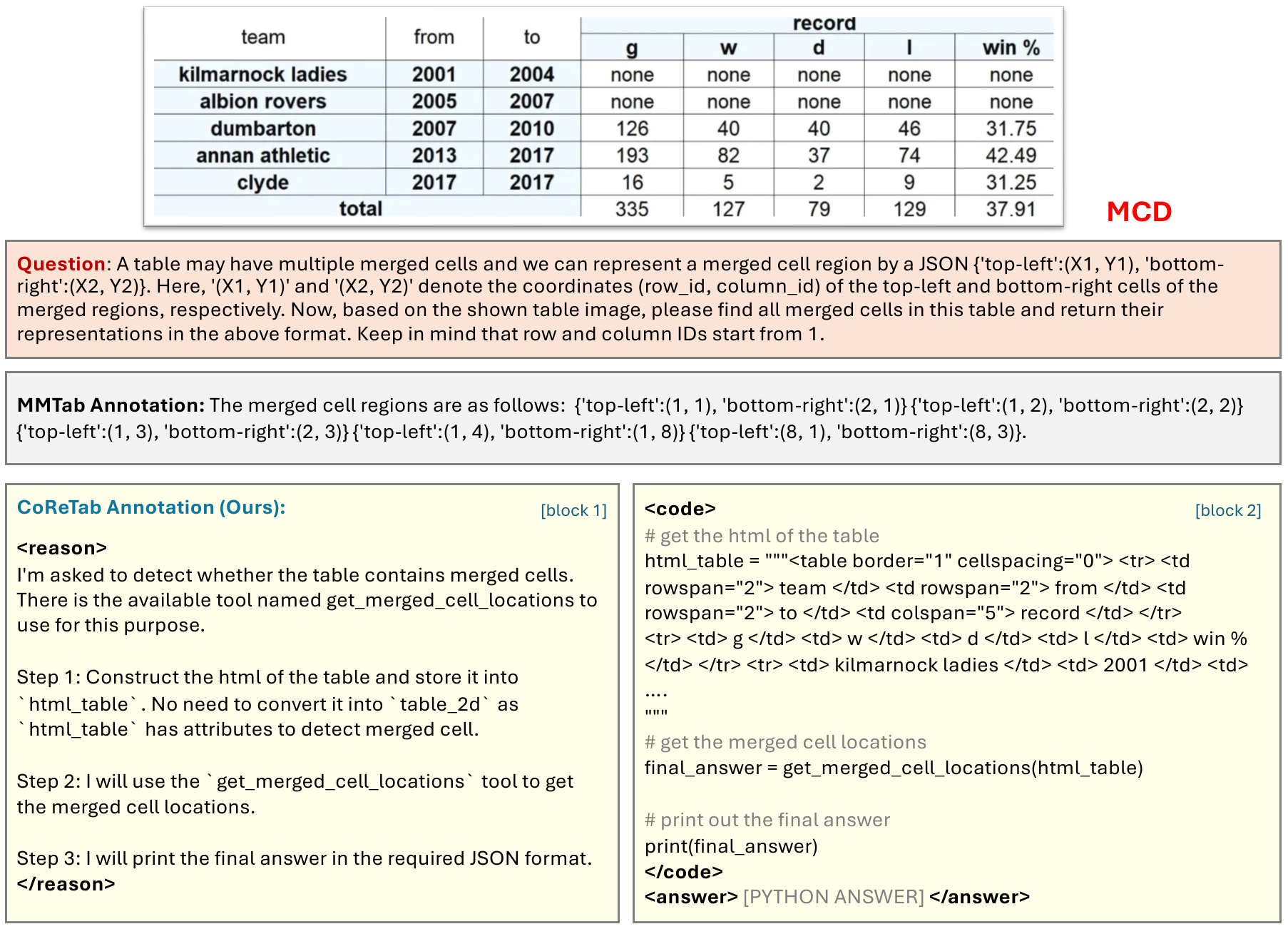}
    \caption{MMTab and \CoReTab annotations for a sample from MCD.}
  \label{fig:extra_example10}
\end{figure*}

\begin{figure*}[ht]\centering
  \includegraphics[width=1.5\columnwidth]{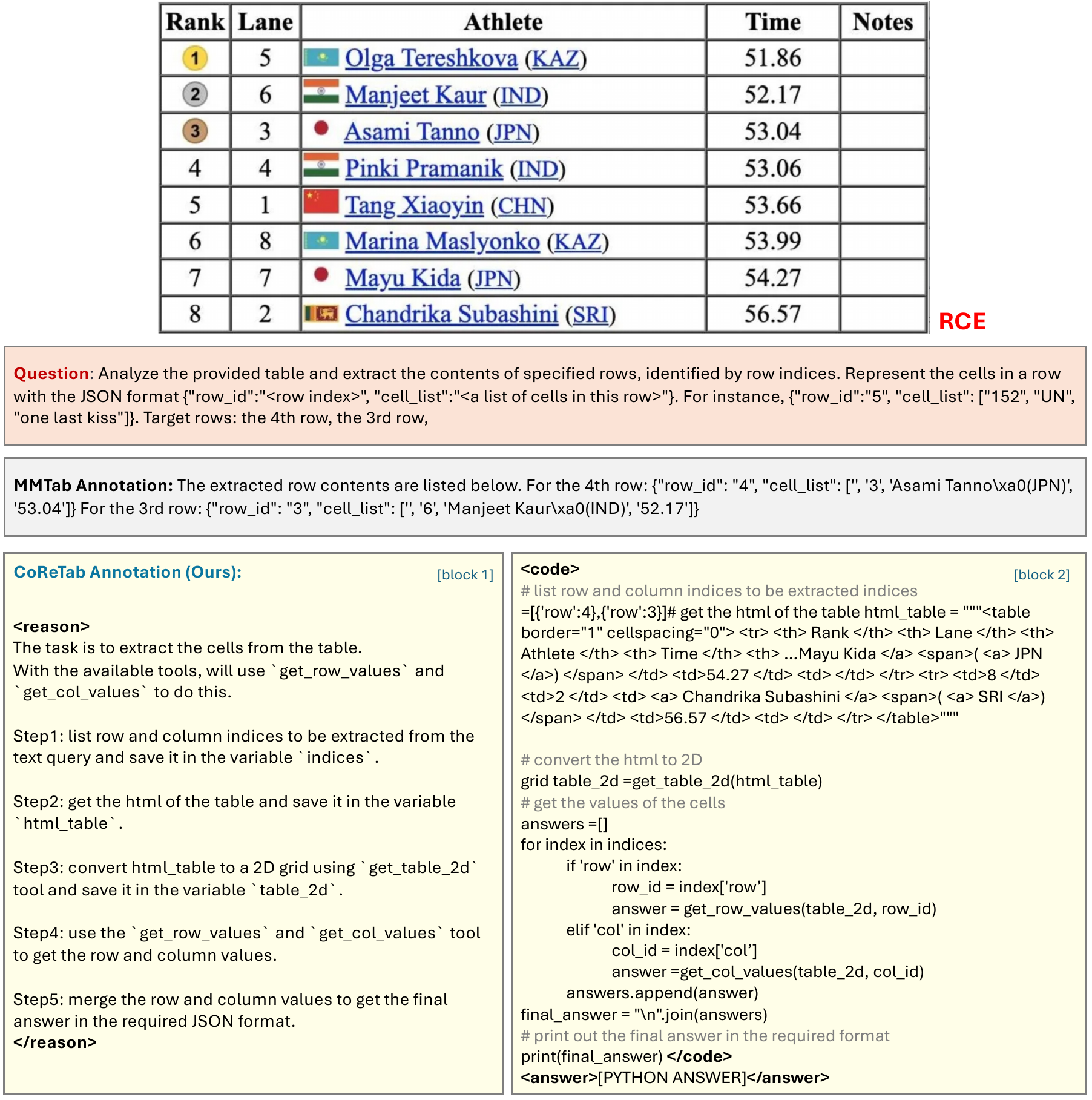}
    \caption{MMTab and \CoReTab annotations for a sample from RCE.}
  \label{fig:extra_example11}
\end{figure*}

We visualize representative samples from MMTab and \CoReTab (Sec.~\ref{sec:dataset}), with examples shown in Fig.~\ref{fig:extra_example1}–\ref{fig:extra_example11}.
Each figure displays the input table image, the user query, and the corresponding MMTab and \CoReTab annotations.

\paragraph{Table Question Answering (TQA).}
Examples from TABMWP, WTQ, HiTab, and TAT-QA are shown in Fig.~\ref{fig:extra_example1}–\ref{fig:extra_example4}.
HiTab and AIT-QA contain large and complex tables, yet MMTab annotations mainly provide short answers.
For numerical reasoning tasks such as TABMWP and TAT-QA, the solutions typically involve multi-step calculations followed by a final answer.
However, MMTab includes only brief answers (short for TAT-QA and a numerical derivation for TABMWP).
In contrast, \CoReTab offers detailed step-by-step reasoning with executable code for all four tasks.

\paragraph{Table Fact Verification (TFV).}
Examples from TabFact, InfoTabs, and PubHealthTab are shown in Fig.~\ref{fig:extra_example4}–\ref{fig:extra_example6}.
These tasks require indexing, querying, and reasoning over the table image to verify a user’s statement.
If the inferred conclusion aligns with the statement, the output is ``correct''; if it contradicts, ``incorrect''; and if uncertain, ``neutral''.
MMTab annotations typically provide only the final answer intentionally, whereas \CoReTab includes explicit reasoning chains and executable code verifying the conclusion.

\paragraph{Table Structure Understanding (TSU).}
Examples from five tasks are shown in Fig.~\ref{fig:extra_example7}–\ref{fig:extra_example11}:
Table Size Detection (TSD) predicts the number of rows and columns;
Table Cell Extraction (TCE) retrieves content by specified indices;
Table Cell Locating (TCL) identifies cell positions matching a query;
Merge Cell Detection (MCD) detects merged regions; and
Row/Column Extraction (RCE) outputs complete rows or columns.
MMTab annotations are of direct answers, while \CoReTab provides annotations with reasoning steps and executable code for verification.

\section{Implementation Details}
\label{more_implementation_details}
\paragraph{The objective in stage 1 and 2.} In the first two stages,
Qwen2.5-VL is trained using the standard cross-entropy loss over both visual and textual tokens. The input to the MLLM is a pair of ⟨table image $I$, text query $Q$⟩, where text query is represented by a sequence of tokens $\mathbf{x} = (x_1, \ldots, x_T)$. The model first encodes the image $I$ into a sequence of visual tokens, which are then concatenated with the text tokens. The objective is to maximize the likelihood of the next token conditioned on the preceding multimodal context:
\begin{equation}
\mathcal{L}_{\text{CE}} = - \sum_{t=1}^{T} \log P_\theta(x_t \mid I, x_{<t}),
\end{equation}
where $P_\theta$ denotes the model’s predicted token distribution parameterized by $\theta$. This objective encourages the MLLM to learn joint reasoning across visual and textual modalities while improving its ability to generate coherent and grounded responses.

\paragraph{The objective in stage 3.}
In stage 3, the model is further refined using RL optimization with GRPO. During training, we sample $G$ rollouts—possible answers $S^1$ to $S^G$ for each ⟨$I$, $Q$⟩ pair. The model is trained to maximize the total reward $\mathcal{R}^{i}$ for each output $S^i$, defined as
\begin{align}
\mathcal{R}^{i} = \mathcal{R}^{i}_{\text{acc}} + \mathcal{R}^{i}_{\text{format}},
\end{align}
where $\mathcal{R}^{i}_{\text{acc}}$ and $\mathcal{R}^{i}_{\text{format}}$ are binary rewards indicating correct answer and correct formatting, respectively, following \citet{guo2025deepseek}. We then compute the relative advantage $\hat{A}^i$ for each output $S^i$ by normalizing its reward with respect to the other outputs in the rollout, as:
\begin{align}
    \hat{A}^i &= \frac{\mathcal{R}^i - \text{mean}(\{\mathcal{R}^1, \dots, \mathcal{R}^G\})}{\text{std}(\{\mathcal{R}^1, \dots, \mathcal{R}^G\})},
    \label{eq:adv}
\end{align}
The overall loss function integrates a clipped surrogate objective with a KL-divergence penalty, formulated as:
\begin{align}
    \mathcal{L}_{\text{GRPO}} = \mathcal{L}_{\text{clip}} - \beta \mathbb{D}_{\text{KL}}[\pi_\theta \| \pi_{\text{ref}}],
    \label{eq:loss-grpo}
\end{align}
where $\mathcal{L}_{\text{clip}}$ adopts the proximal policy optimization mechanism. This is to constrain policy updates to avoid excessively large parameter shifts. The KL divergence term $\mathbb{D}_{\text{KL}}$ further regularizes the policy $\pi_\theta$ to stay close to the reference model $\pi_{\text{ref}}$. Refer to \citet{grpo, guo2025deepseek} for more details.

\paragraph{Implementation Settings.}
All experiments are conducted on 8 NVIDIA A200 80GB Tensor Core GPUs using DeepSpeed \cite{rajbhandari2020zero,rasley2020deepspeed} with ZeRO optimization \cite{rajbhandari2020zero} and the Swift framework~\cite{zhao2025swift} for training. 
We employ the AdamW optimizer across all three stages, training each for two epochs with a per-GPU batch size of 1. Gradient accumulation is used to achieve a global batch size of 256 for parameter updates. During each training stage, we applied LoRA to the MLLM model with a rank of 8 and a scaling factor (\texttt{lora\_alpha}) of 32, targeting all linear layers and specific projections (\texttt{o\_proj}, \texttt{up\_proj}, \texttt{v\_proj}, \texttt{q\_proj}, \texttt{gate\_proj}, \texttt{k\_proj}, \texttt{down\_proj}) in the language model, while keeping the vision encoder frozen. In stages 1 and 2, the learning rate is set to $2\times10^{-5}$, with a weight decay of 0.1, $\beta_1{=}0.9$, and $\beta_2{=}0.95$.

During stage 3 with GRPO, we set the number of groups to $G{=}4$ (i.e., four rollouts per ⟨$I$, $Q$⟩ pair) and use a sampling temperature of 0.5 to encourage diverse reasoning paths. The maximum sequence length is set to $L{=}4096$ to allow the model to generate complete responses containing both natural language reasoning and executable Python code. The KL-divergence coefficient $\beta$ in Eq.~\ref{eq:loss-grpo} is fixed at 0.05, and the learning rate for the policy model is $2\times10^{-6}$.

During inference, we employ the vLLM engine \cite{kwon2023efficient} for efficient decoding. To ensure deterministic and reproducible outputs, we set the sampling temperature to \texttt{0}, corresponding to greedy decoding. This configuration eliminates randomness in token selection, allowing consistent evaluation across all samples.

\subsection{System Prompt to \CoReTab-Qwen2.5-VL}
The system prompt is prepended to the \CoReTab-Qwen2.5-VL model during both training and inference, differing slightly from the one used for the LLM annotator. For TSU tasks, the descriptions of pre-defined functions are optionally appended to the system prompt before being passed to Qwen2.5-VL. The input to the MLLM is structured as follows:

\begin{tcolorbox}[
  title={\footnotesize \centering Structured input to \CoReTab-Qwen2.5-VL},
  colback=white,
  colframe=navy,
  colbacktitle=navy,
  coltitle=white,
  fonttitle=\bfseries,
  arc=1.5mm,      
  boxrule=1.0pt
]\footnotesize
    You are an expert in table analysis. Your task is to process a table image to answer a question with a code-driven reasoning explanation. \\[0.6em]
    Your output must strictly follow this structure:\\[0.4em]
    1. Reasoning — enclosed in <reason> tags. Provide clear, concise, and verifiable reasoning steps that logically lead to the final answer. \\[0.4em]
    2. Python Code — enclosed in <code> tags. Write clean, executable Python code that computes the final answer based on the reasoning. The code must print the answer in the correct format. \\[0.4em]
    3. Final Answer — enclosed in <answer> tags. This must contain the final answer or the printed output of the code snippet. \\[0.4em]
  {\color{gray}\textbf{Tool Use}: \{\{ \texttt{Tool Use Descriptions} \}\} \# optional} \\[0.4em]
  \textbf{Input}: \{\{ \texttt{table image, question} \}\}
\end{tcolorbox}    

\section{Additional Qualitative Results}

\begin{figure*}[ht]\centering
  \includegraphics[width=1.8\columnwidth]{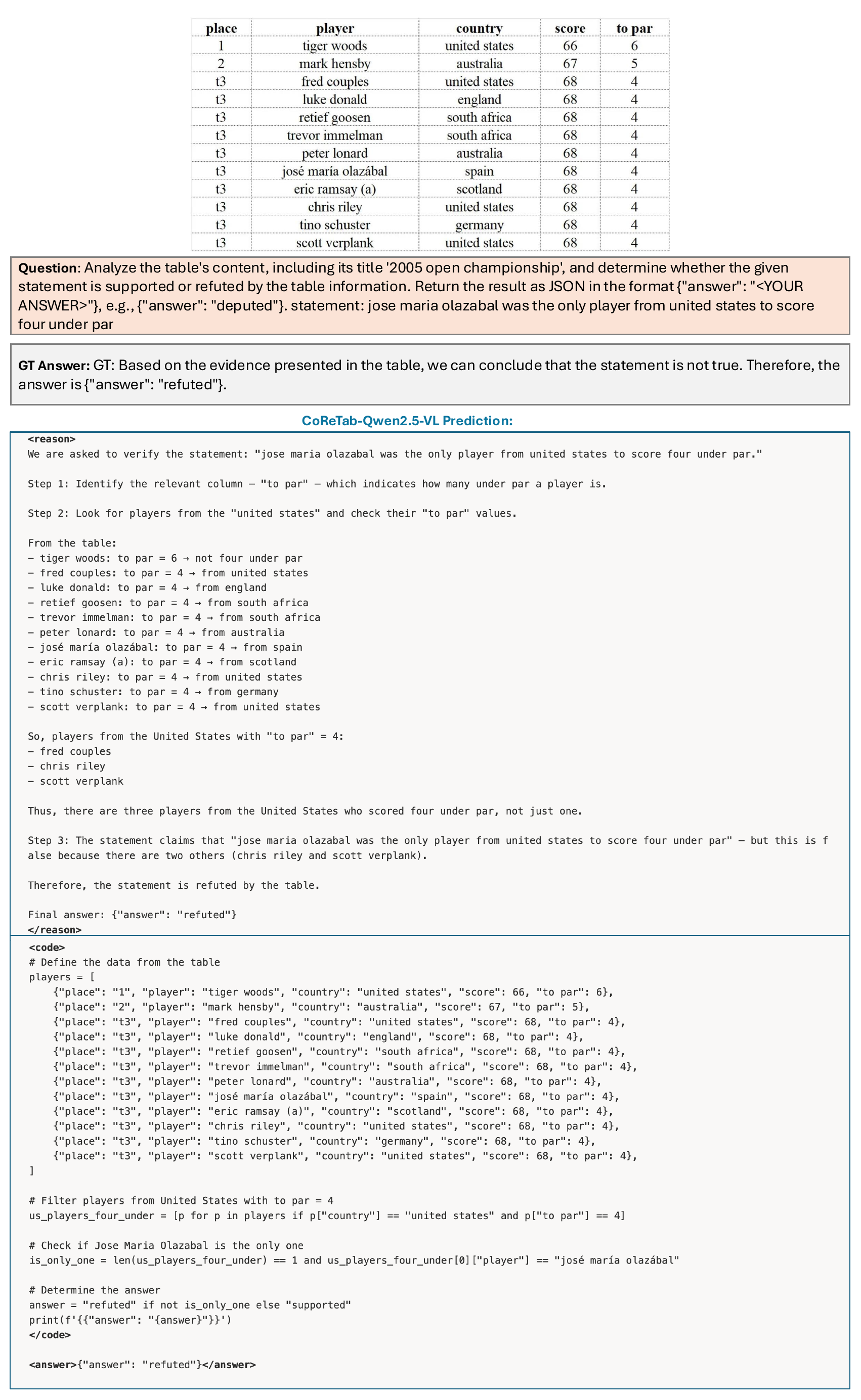}
    \caption{Generated response of \CoReTab-Qwen2.5-VL on the test sample.}
  \label{fig:extra_prediction1}
\end{figure*}

\begin{figure*}[ht]\centering
  \includegraphics[width=2\columnwidth]{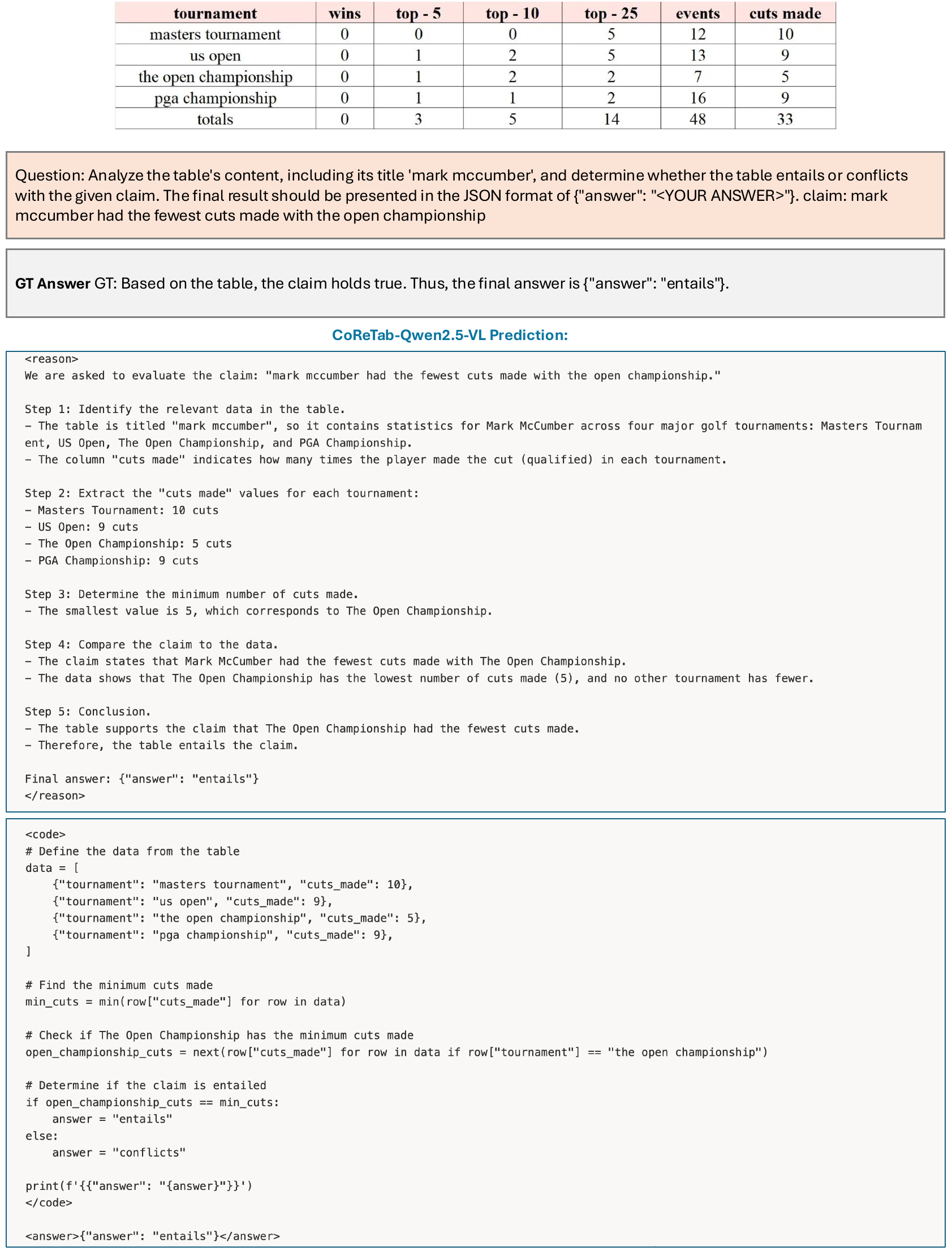}
    \caption{Generated response of \CoReTab-Qwen2.5-VL on the test sample.}
  \label{fig:extra_prediction2}
\end{figure*}

\begin{figure*}[ht]\centering
  \includegraphics[width=2.0\columnwidth]{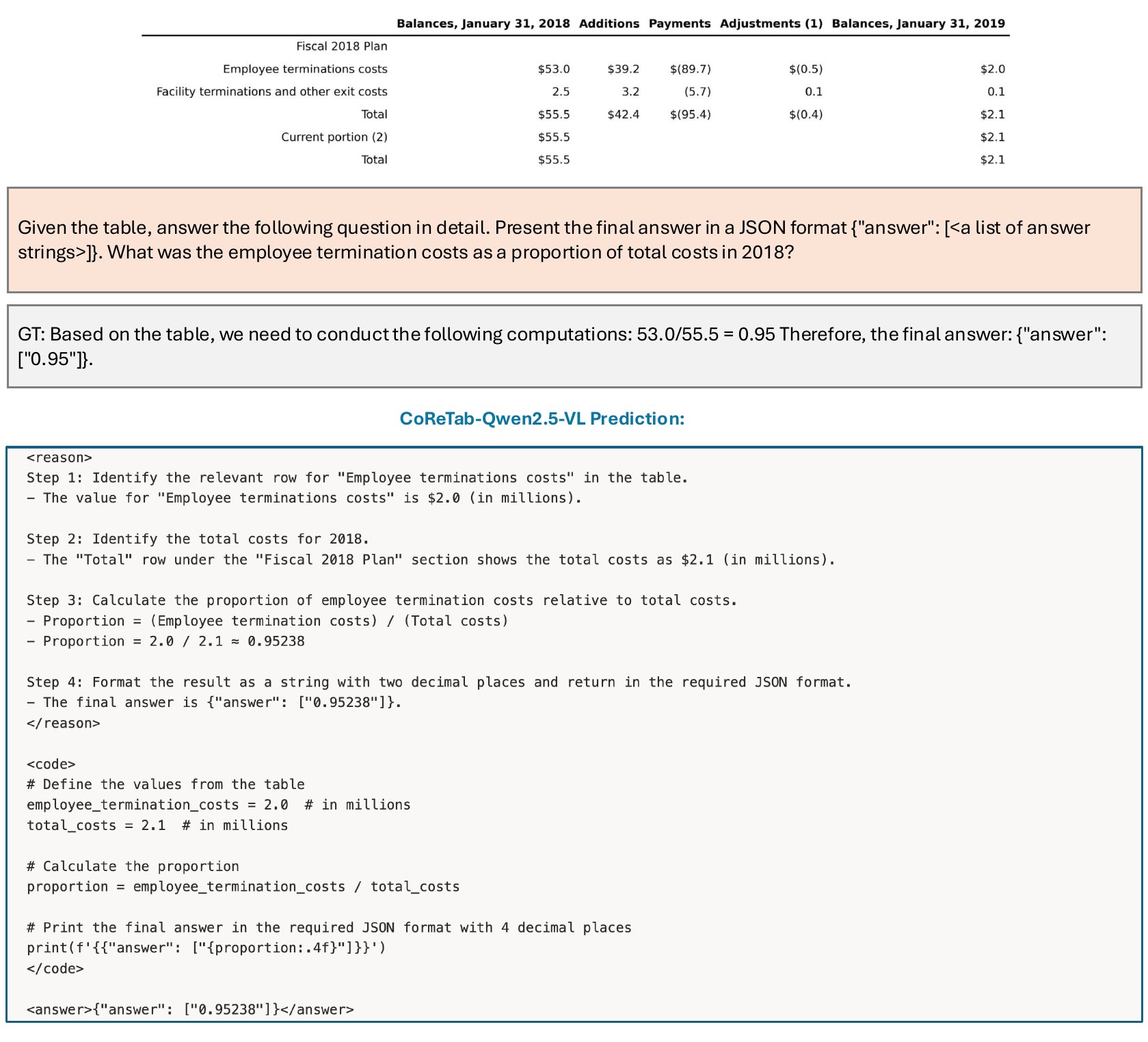}
    \caption{Generated response of \CoReTab-Qwen2.5-VL on the test sample.}
  \label{fig:extra_prediction3}
\end{figure*}

\begin{figure*}[ht]\centering
  \includegraphics[width=2.\columnwidth]{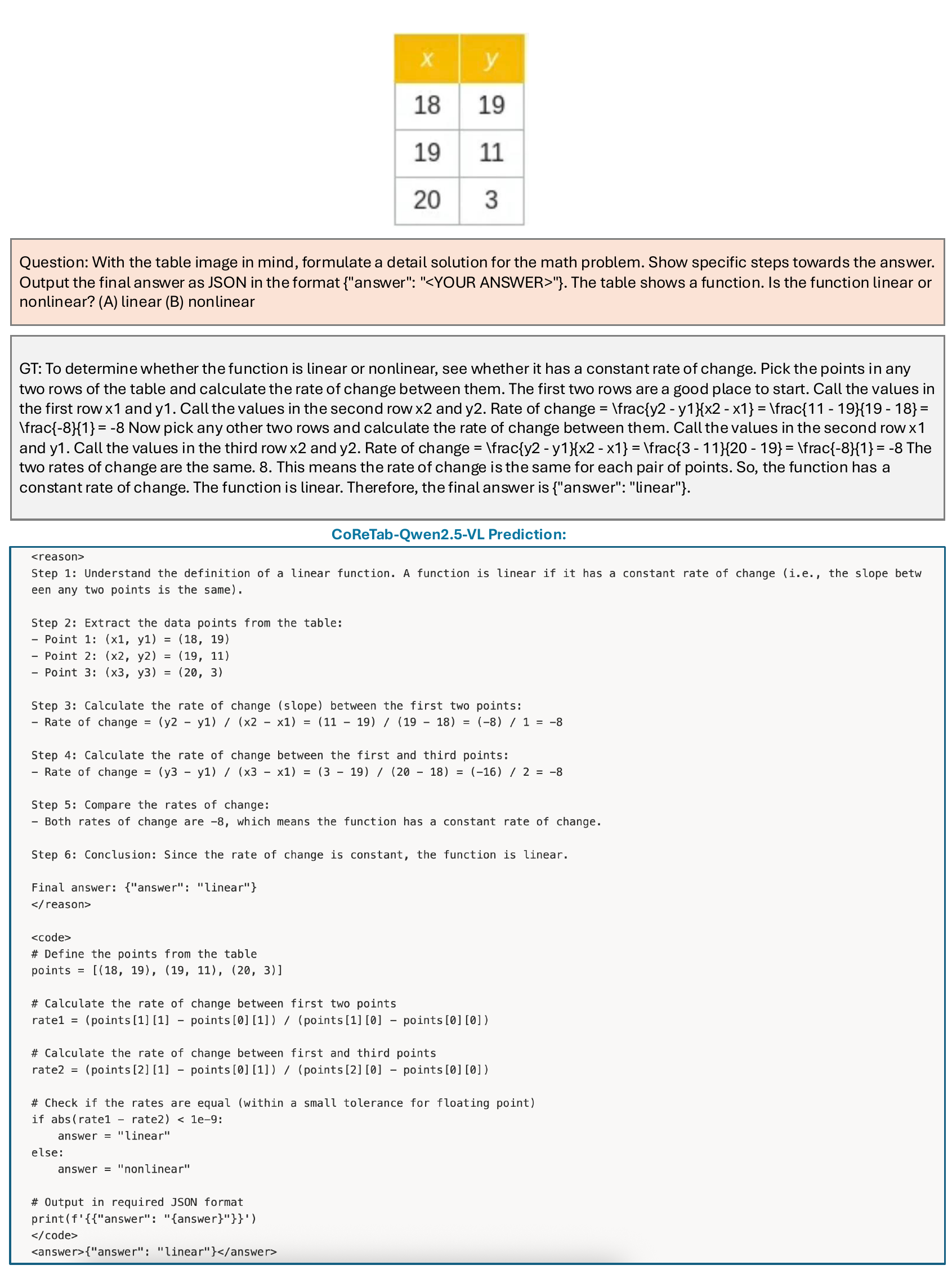}
    \caption{Generated response of \CoReTab-Qwen2.5-VL on the test sample.}
  \label{fig:extra_prediction4}
\end{figure*}

We provide additional results from \CoReTab-Qwen2.5-VL on the test samples in Fig.~\ref{fig:extra_prediction1}–\ref{fig:extra_prediction4}. The results show that the model generates correct answers along with detailed step-by-step reasoning and executable Python code that can be verified.

\end{document}